\documentclass[10pt,twocolumn,letterpaper]{article}

\usepackage[pagenumbers]{cvpr} 
\usepackage{overpic}
\usepackage{adjustbox}
\usepackage{colortbl}

\usepackage{etoc}
\etocdepthtag.toc{mtchapter}
\etocsettagdepth{mtchapter}{subsection}
\etocsettagdepth{mtappendix}{none}

\definecolor{cvprblue}{rgb}{0.21,0.49,0.74}
\usepackage[pagebackref,breaklinks,colorlinks,allcolors=cvprblue]
{hyperref}
\usepackage{multirow}
\usepackage{pifont}

\def\paperID{2019} 
\def\confName{CVPR}
\def\confYear{2025}

\newtheorem{assumption}{Assumption}
\newcommand{\methodname}{\emph{HumanMM}\xspace}
\newcommand{\dataname}{\emph{ms}-Motion\xspace}
\newcommand{\msaist}{\emph{ms}-AIST\xspace}
\newcommand{\msm}{\emph{ms}-H3.6M\xspace}
\newcommand{\mshmr}{\emph{ms-HMR}\xspace}
\title{HumanMM: Global Human Motion Recovery from Multi-shot Videos}
\author{Yuhong Zhang$^{1,2,\ddag}$\thanks{Equal contribution, \ $^\ddag$Core contributor, \ $^\dag$Corresponding author. Work done by Yuhong Zhang, Guanlin Wu, Ling-Hao Chen, Jing Lin and Jiamin Wu during the internship at IDEA Research.} \quad
Guanlin Wu$^{2,3, \ddag*}$ \hfill
Ling-Hao Chen$^{1,2,\ddag}$ \hfill
Zhuokai Zhao$^{4}$ \hfill
Jing Lin$^{1,2}$ \\
Xiaoke Jiang$^{2}$ \ \hfill
Jiamin Wu$^{2,5}$ \  \hfill
Zhuoheng Li$^{6}$ \  \hfill
Hao Frank Yang$^{3}$ \  \hfill
Haoqian Wang$^{1\dag}$ \  \hfill
Lei Zhang$^{2\dag}$ \\
{\small $^1$Tsinghua University \hfill
$^2$IDEA Research  \hfill
$^3$Johns Hopkins University  \hfill
$^4$University of Chicago  \hfill
$^5$HKUST  \hfill
$^6$HKU} \\
{\texttt{\small \{dsyuhong, guanlinwu0930, thu.lhchen\}@gmail.com}}\\
Project page: \url{https://zhangyuhong01.github.io/HumanMM}
}

\makeatletter
\apptocmd\@maketitle{{\myfigure{}\par}}{}{}
\makeatother

\begin{document}
\definecolor{zhendong}{HTML}{c1a9a4}
\definecolor{malong}{HTML}{59d7cb}

\newcommand\myfigure{%
\vspace{-2em}
\centering
   \includegraphics[width=\linewidth]{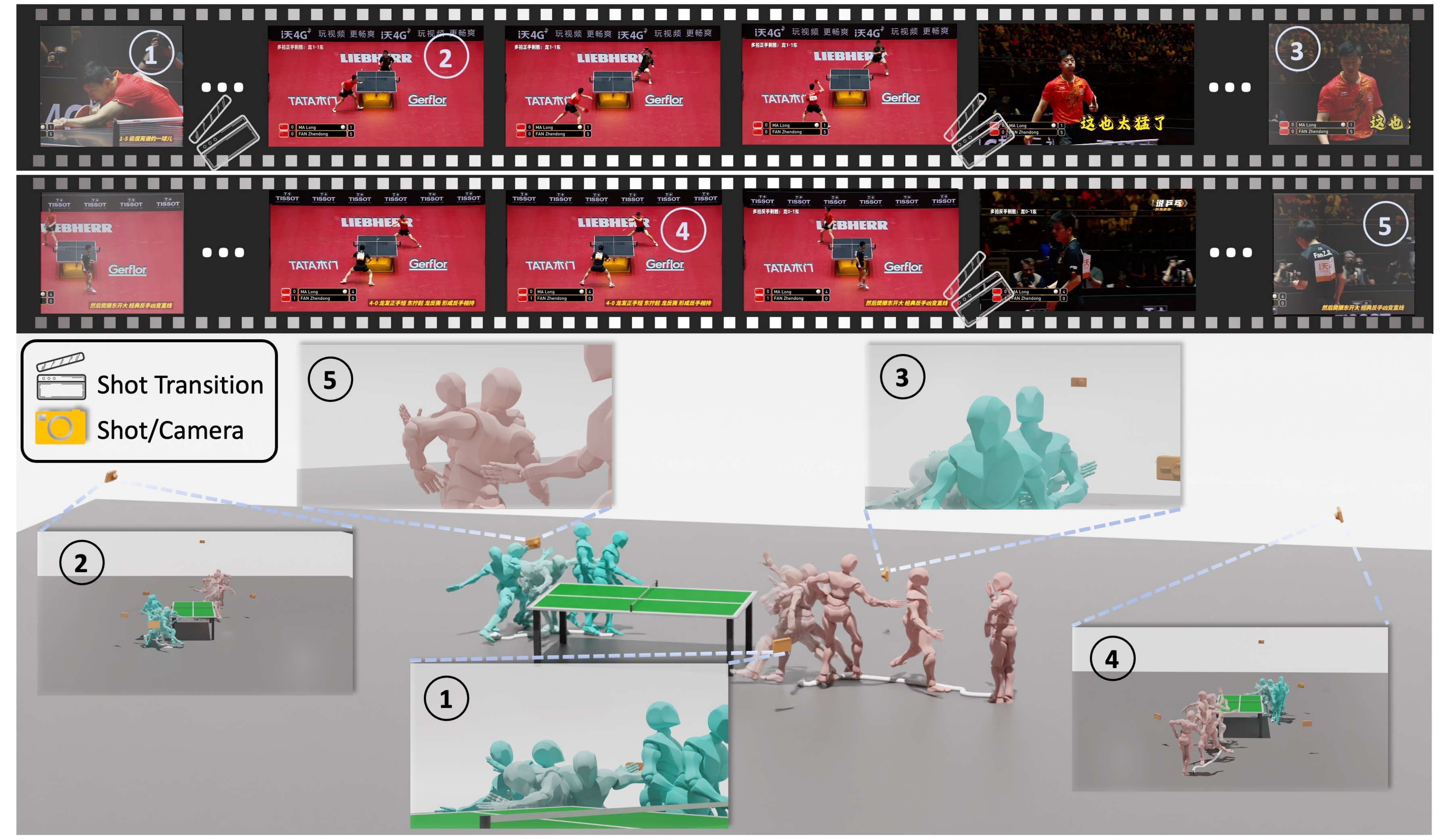}
\vspace{-2em}
\captionof{figure}{{\bf Recovering a human motion from multi-shot videos.} \textbf{Top}: We take two multi-shot table tennis game videos with shot transitions as input. We aim to recover two motions of two athletes (Long MA and Zhendong FAN) from two videos, respectively. The first video is recorded by three shots (``\ding{172}'', ``\ding{173}'', and ``\ding{174}'' ), and the second one is recovered by two shots  (``\ding{175}'' and ``\ding{176}'' ). \textbf{Bottom}: We recover two motions (\textcolor{malong}{Long MA} in green and \textcolor{zhendong}{Zhendong FAN} in pink), different shots, and camera poses for each multi-shot video. The recovered motion is aligned with the motion in the videos.}
\label{fig:teaser}
\vspace{0.5em}
}

\maketitle
\begin{abstract}
In this paper, we present a novel framework designed to reconstruct long-sequence 3D human motion in the world coordinates from in-the-wild videos with multiple shot transitions. Such long-sequence in-the-wild motions are highly valuable to applications such as motion generation and motion understanding, but are of great challenge to be recovered due to abrupt shot transitions, partial occlusions, and dynamic backgrounds presented in such videos. Existing methods primarily focus on single-shot videos, where continuity is maintained within a single camera view, or simplify multi-shot alignment in camera space only. In this work, we tackle the challenges by integrating an enhanced camera pose estimation with Human Motion Recovery (HMR) by incorporating a shot transition detector and a robust alignment module for accurate pose and orientation continuity across shots. By leveraging a custom motion integrator, we effectively mitigate the problem of foot sliding and ensure temporal consistency in human pose. Extensive evaluations on our created multi-shot dataset from public 3D human datasets demonstrate the robustness of our method in reconstructing realistic human motion in world coordinates. 
\vspace{-2em}

\end{abstract}    
\section{Introduction}
\label{sec:intro}

In recent years, significant advances have been made in 3D human pose estimation, particularly in enhancing the accuracy of human motion recovery (HMR)\footnote{In this paper, the ``human mesh recovery'' refers to recovery in the camera coordinates and the ``human motion recovery'' denotes recovery in the world coordinates. Unless specified otherwise, HMR refers to \textbf{human motion recovery}.} from monocular video sequences.
HMR has demonstrated extensive applications in areas such as human-AI interaction~\cite{wang2022towards, xiao2024unified}, human motion understanding~\cite{tm2t,motiongpt,pan2023synthesizing, wang2023learning}, and motion generation~\cite{motionclip, temos, tm2t, motiondiffuse, mdm, humanise, mld, physdiff, t2mgpt, diffprior, gmd, motionclr, unihsi, omnicontrol,humantomato,motionlcm,humanmac,lu2024scamo,motionlcmv2}.
While existing methods~\citep{goel2023humans} have achieved relatively high performance in recovering human mesh in camera coordinates, estimating human motion in world coordinates remains challenging~\citep{ye2023decoupling, wang2024tram, shin2024wham, shen2024gvhmr} due to inaccurate camera pose estimation and the complexity of reconstructing human motion spatially.

Most current progress in 3D human motion community mainly benefits from large scale data~\citep{angjoo2019learning, goel2023humans, wang2024tram, shin2024wham, shen2024gvhmr, kocabas2024pace}, and long-sequence videos.
These resources enhance estimation accuracy for HMR methods and improve the understanding and generation of longer motion sequences for tasks such as motion understanding~\citep{chen2024motionllm, plappert2018learning, tm2t} and generation~\citep{motionclip, temos, avatarclip, tm2t, motiondiffuse, teach, mdm, humanise, mld, mofusion, physdiff, t2mgpt, diffprior, remodiffuse, gmd, motionclr, unihsi, omnicontrol, humantomato, tlcontrol, momask, promotion, amd, b2ahdm, emdm, stmc, hu2025mona, move,qing2023storytomotion}, even when annotations are derived from markerless capturing methods like pseudo labels~\citep{pang2022benchmarking, moon2022neuralannot, moon2023three, yi2023generating}. 

A promising approach to enlarge the scale of the motion databases is to estimate human motions from \textit{unlimited} online videos in a \textit{markerless} manner. However, many long-sequence online videos are recorded with multiple shots, referred to as multi-shot videos\footnote{In this paper, a \textbf{multi-shot video} refers to a long-sequence video containing multiple shot transitions. We assume that the camera intrinsics remain consistent across different shots within a multi-shot video.}, especially prevalent in domains such as sports broadcasting, talk shows, and concerts. In filmmaking and television live show, a ``shot'' denotes an individual camera view capturing a specific moment or action from a particular vantage point~\citep{bowen2013grammar}.

\begin{figure}
    \centering
    \includegraphics[width=0.95\linewidth]{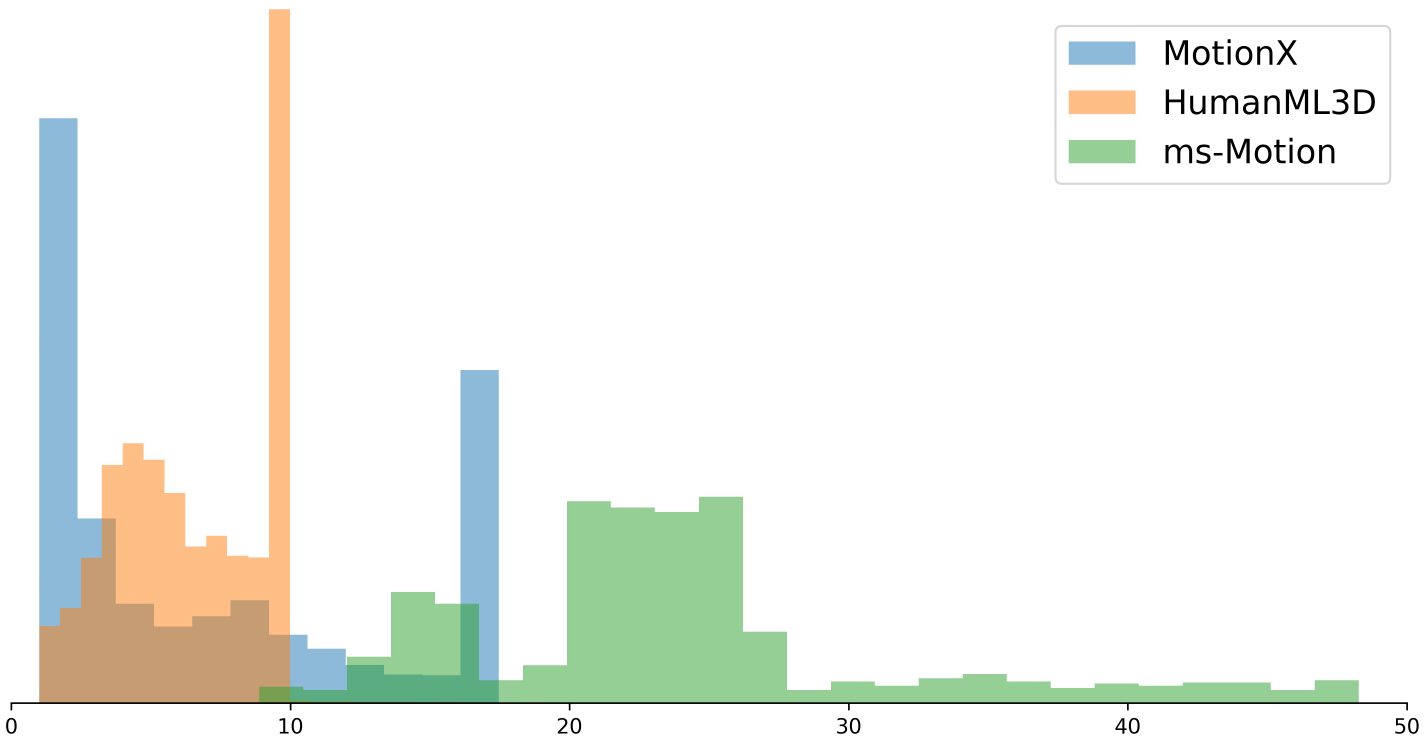}
    \vspace{-1em}
    \caption{The comparison between the distribution of sequence lengths in different existing large-scale markerless motion datasets with ours. The $x$-axis and $y$-axis denote the duration time (s) and percentage of video number, respectively. Our dataset (in green) contains more portion of long-sequence videos in general.}
    \vspace{-1em}
    \label{fig:frame_distribution}
\end{figure}

Segmenting multi-shot videos into separate shots inevitably reduces the length of the video sequences, which can be detrimental to tasks that benefit from longer sequences, such as long motion generation~\cite{petrovich2024multitrack,qing2023storytomotion}. This limitation is highlighted in the existing datasets~\citep{lin2023motionx,humanml3d}, where the longest clip is less than 20 seconds after segmentation, as shown in \cref{fig:frame_distribution}. Moreover, focusing exclusively on online single-shot videos diminishes the utilization ratio of available online videos and may negatively impact the diversity of scenarios represented in the created datasets.

Therefore, \textit{how to address the issue of discontinuities caused by shot transitions} is notoriously difficult in the community. To resolve this problem, previous works~\citep{pavlakos2022multishot, wu2023clipfusion, wang2022nemo, baradel2021leveraging} have proposed algorithms to address human mesh recovery in a camera space from movies containing shot change between long shots and close-ups.

However, recovering human motions in world coordinates from multi-shot videos presents two fundamental challenges that remain underexplored.
1) \textit{How to align the human motion and orientation in the world coordinates during shot transitions?} Ensuring continuity of human orientation and pose across shots is complicated by factors such as partial visibility of human body (\eg transitioning from long shot to close-up) and changes in human orientation (\eg two long shots from different viewpoints). 
These issues, caused by abrupt changes in camera viewpoints, necessitate robust alignment mechanisms.
%
2) \textit{How to reconstruct accurate human motion in world coordinates?} Existing approaches employ Simultaneous Localization and Mapping (SLAM) methods to estimate camera parameters, which are then used to project recovered human meshes from camera to world coordinates~\citep{ye2023decoupling, shin2024wham, wang2024tram, shen2024gvhmr}. 
This process requires highly accurate camera estimation and must address motion consistency and foot sliding in the recovered human motion within the world space.

Despite these challenges, human motion in multi-shot videos often remain continuous across shots, even as camera viewpoints change. 
This observation suggests that with appropriate handling of shot transitions and camera motion, it is possible to reconstruct consistent and complete 3D human motions throughout multi-shot videos.

In this paper, we propose a novel framework \methodname, \underline{Human} \underline{M}otion recovery from \underline{M}ulti-shot videos, to address these challenges. It integrates human pose estimation across shots with robust camera estimation in the world space. 
Firstly, we develop a shot transition detector to identify frames with shot transitions.
To ensure a more robust camera pose estimation, we introduce an enhanced SLAM method incorporating long-term tracking of feature points and exclusion of moving human from bundle adjustment process.
We utilize existing HMR method integrated with our enhanced camera estimation to get the initial human parameters for each separated shot.
Subsequently, we implement an alignment module to align human orientation based on stereo calibration and smooth human poses through a trained multi-shot HMR encoder, which effectively captures the temporal context of human movements across different shots. 
Finally, after aligning human and camera parameters between shot transitions, we train a motion decoder and a trajectory refiner to smooth the human pose and mitigate issues such as foot sliding, thereby enhancing the overall motion consistency in the reconstructed 3D human motions.

Our contributions can be summarized as follows.
\begin{itemize}
    \item We present the first approach to reconstruct human motion from multi-shot videos in world coordinates.
    
    \item We introduce \methodname, a HMR framework for multi-shot videos. It includes an enhanced camera trajectory estimation method, a human motion alignment module and a motion integrator to ensure accurate and consistent recovery of human pose and orientation in world coordinates across different shots in the whole video.

    \item We develop a multi-shot video dataset \dataname to evaluate the performance of HMR from multi-shot videos, based on existing public datasets such as AIST~\citep{li2021aistpp} and Human3.6M~\citep{ionescu2014human36m}. Extensive experiments on related benchmarks verify the effectiveness of our method.
\end{itemize}

\section{Related Work}
\label{sec:related_work}


\subsection{HMR from One-shot Video}


One-shot videos, captured with a single camera without shot transitions, has been extensively studied within the community for human mesh and motion recovery.

\noindent\textbf{Human mesh recovery in camera coordinates} can be broadly categorized into two approaches: optimization-based methods~\cite{romero2017mano, bogo2016smplify, arnab2019exploiting, osman2020star, huang2017towards} and regression-based methods~\cite{kanazawa2017end, kocabas2019vibe, kolotouros2019cmr, angjoo2019learning, luo20203dhuman}. 
With the significant advancements of transformer~\cite{vaswani2023attention}, HMR2.0~\cite{goel2023humans} has surpassed previous methods and benefits several downstream tasks related to HMR.

Although there are several previous works tried to recover motions in world coordinates with multi-camera capture system~\cite{huang2022rich, li2021aistpp} and IMU-based methods~\cite{kaufmann2023emdb, yi2021transpose} and enjoy relatively satisfying results, this setup limits their use for applications of \textit{infinite} in-the-wild monocular videos.
To address this limitation, several attempts~\cite{ye2023decoupling, shin2024wham, wang2024tram, shen2024gvhmr} integrate SLAM into the HMR pipeline by first estimating the camera pose using SLAM methods, \eg DROID-SLAM~\cite{teed2021droid} or DPVO~\cite{teed2023deep}, and then project the recovered human motion from camera to world coordinates. 
To exclude the inconsistencies caused by dynamic objects, such as moving humans, TRAM~\cite{wang2024tram} modifies DROID-SLAM by incorporating human masking and depth-based distance rescaling. 
However, DROID-SLAM performs dense bundle adjustment (DBA) on feature maps from downsampled images and selects features based only on two consecutive frames rather than long-term video sequences~\cite{teed2021droid, teed2023deep, chen2024leap}. Consequently, masking significantly reduces the number of informative and consistent features, especially when humans occupy large portions of the image, leading to inaccuracies.
Therefore, developing a SLAM method that retains sufficient and representative features for DBA after masking is important. 
%

\subsection{HMR from Multi-shot Video}

Multiple shots are fundamental elements of cinematic storytelling and live performances, utilizing various camera positions and focal lengths to create immersive and detailed viewing experiences for audiences~\cite{bowen2013grammar}.
However, most marker-based motion capture (MoCap) datasets~\citep{kaufmann2023emdb, li2021aistpp, huang2022rich, mahmood2019amass, von20183dpw} consist single-shot videos only, resulting in limited research on HMR from multi-shot videos.

\vspace{-0.1em}
Recovering human motion from multi-shot videos in camera coordinates is already challenging. This is because treating each pose estimation result of each shot separately leads to inconsistencies when combining all estimations, caused by partially or fully invisible human bodies across shot transitions.
Pavlakos \etal~\citep{pavlakos2022multishot} addresses this issue by focusing on shot changes from long shots to close-ups, which are common in film. They develop smoothness constraints within a temporal Human Mesh and Motion Recovery (t-HMMR) model to infer motions during occlusions caused by shot transitions.

Advancements in HMR methods~\cite{shen2024gvhmr} for single-shot videos in world coordinates have paved the way for extending HMR to multi-shot videos with varying camera viewpoints. 
However, aligning human orientation, body pose, and translation continuously across multi-shot videos in world coordinates underexplored. Effective alignment is crucial to maintain motion continuity and coherence, especially when dealing with diverse camera perspectives and abrupt transitions between shots.

\vspace{-0.1em}

In summary, while substantial progress has been made in HMR from single-shot videos, extending these techniques to multi-shot videos requires addressing additional complexities related to camera pose alignment and motion consistency across shot transitions. We address this challenge by proposing a novel pipeline that ensures accurate and continuous 3D HMR from multi-shot monocular videos.

\section{Method}
\label{sec:method}

\definecolor{number1}{HTML}{2f8d88}
\definecolor{number2}{HTML}{6b339d}
\definecolor{number3}{HTML}{5d8c3d}

\begin{figure*}[t]
    \centering
    \begin{overpic}[width=0.995\linewidth]{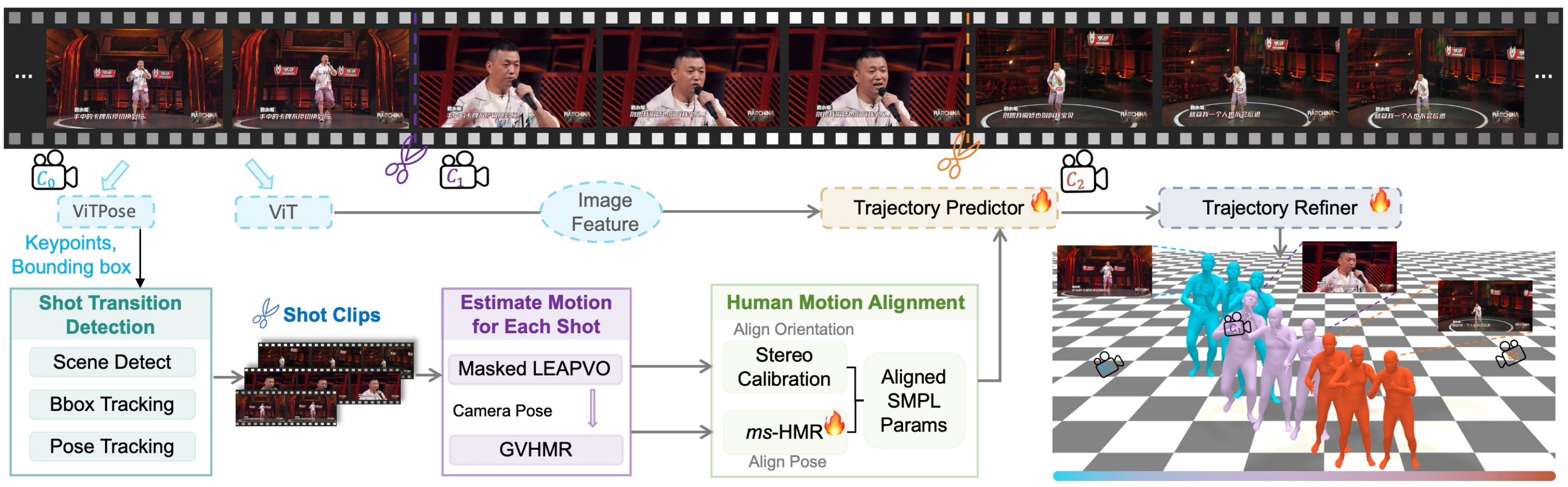}
        \put(35.5, 4.4){\scriptsize $\mathbf{G}$}
        \put(41, 6.4){\scriptsize $\Gamma$}
        \put(40.7, 8.0){\footnotesize $\mathbf{R}, \mathbf{T}$}
        \put(40.5, 3.9){\scriptsize $\theta, \beta, \tau$}
        \put(55.2, 1.2){\footnotesize $\phi, \beta, \Gamma, \tau$}
    \end{overpic}
    \vspace{-0.6em}
    \caption{\textbf{The overview of \methodname}. \methodname processes multi-shot video sequences by first extracting motion feature such as keypoints and bounding boxes, using ViTPose~\cite{xu2022vitpose} and image feature using ViT~\cite{dosovitskiy2020vit}. These features are then segmented into single-shot clips via \textcolor{number1}{\textit{Shot Transition Detection}} (\cref{sec:shot_detector}). Initialized camera (camera rotation $\mathbf{R}$ and camera translation $\mathbf{T}$) and human (SMPL) parameters for each shot are estimated using \textcolor{number2}{\textit{Masked LEAP-VO}} (\cref{sec:initialization}) and \textcolor{number2}{GVHMR}~\cite{shen2024gvhmr}. Human orientation is aligned across shots through \textcolor{number3}{\textit{camera calibration}} (\ref{sec:orientation_alignment}), and \textcolor{number3}{\mshmr} (\cref{sec:pose_alignment}) ensures consistent pose alignment. Finally, a bi-directional \textit{LSTM-based trajectory predictor} with \textit{trajectory refiner} predicts trajectory based on aligned motion and mitigates foot sliding throughout the video.}
    \label{fig:pipeline}
\end{figure*}

In this section, we propose \methodname to recover human motion from multi-shot videos. The system overview is shown in~\cref{fig:pipeline}. 
Given an input video sequence $\mathbf{V} = \{I_t\}_{t=1}^{T}$ of length $T$, where $I_t$ denotes the $t$-th frame, our objective is to recover human motion in world coordinates. 
We begin by detecting shot transition frames based on human bounding box (\aka bbox) and 2D keypoints (\aka KPTs) through a \textit{shot transition detector}~(\cref{sec:shot_detector}).
For each clipped shot, we initialize the camera pose (camera rotation and camera translation) and recover initial human motion in world coordinates~(\cref{sec:initialization}).
The initialized SMPL parameters and camera poses are then fed into a \textit{human motion alignment} module~(\cref{sec:alignment}), which aligns human orientations via camera calibration based on human 2D KPTs and smooth the human pose by incorporating pose information across different shots.
Additionally, it refines the entire motion sequence through whole video using a temporal motion encoder \textit{ms}-HMR.
Finally, we introduce a post-processing module for motion integration~(\cref{sec:network}).
%
%

\subsection{Preliminary: 3D Human Model}
\label{sec:preliminary}
%


Our method aims to recover motions in world coordinates in the SMPL~\citep{loper2015smpl} format, whose pose at frame $t$ can be represented as $\mathcal{M}_t(\theta_t, \beta_t, \Gamma_t, \tau_t) \in \mathbb{R}^{6890 \times 3}$. Here, the body pose, body shape, root orientation, and translation are $\theta_t \in \mathbb{R}^{23 \times 3}$, $\beta_t \in \mathbb{R}^{10}$, $\Gamma_t \in \mathbb{R}^{3}$, and $\tau_t \in \mathbb{R}^{3}$, respectively. We use $\mathbf{K}^{2D}_t$ to denote human 2D KPTs at each frame $t$.

\subsection{Shot Transition Detector For Multi-shot Video}

\label{sec:shot_detector}

\begin{figure}[!t]
    \centering
    \includegraphics[width=0.99\linewidth]{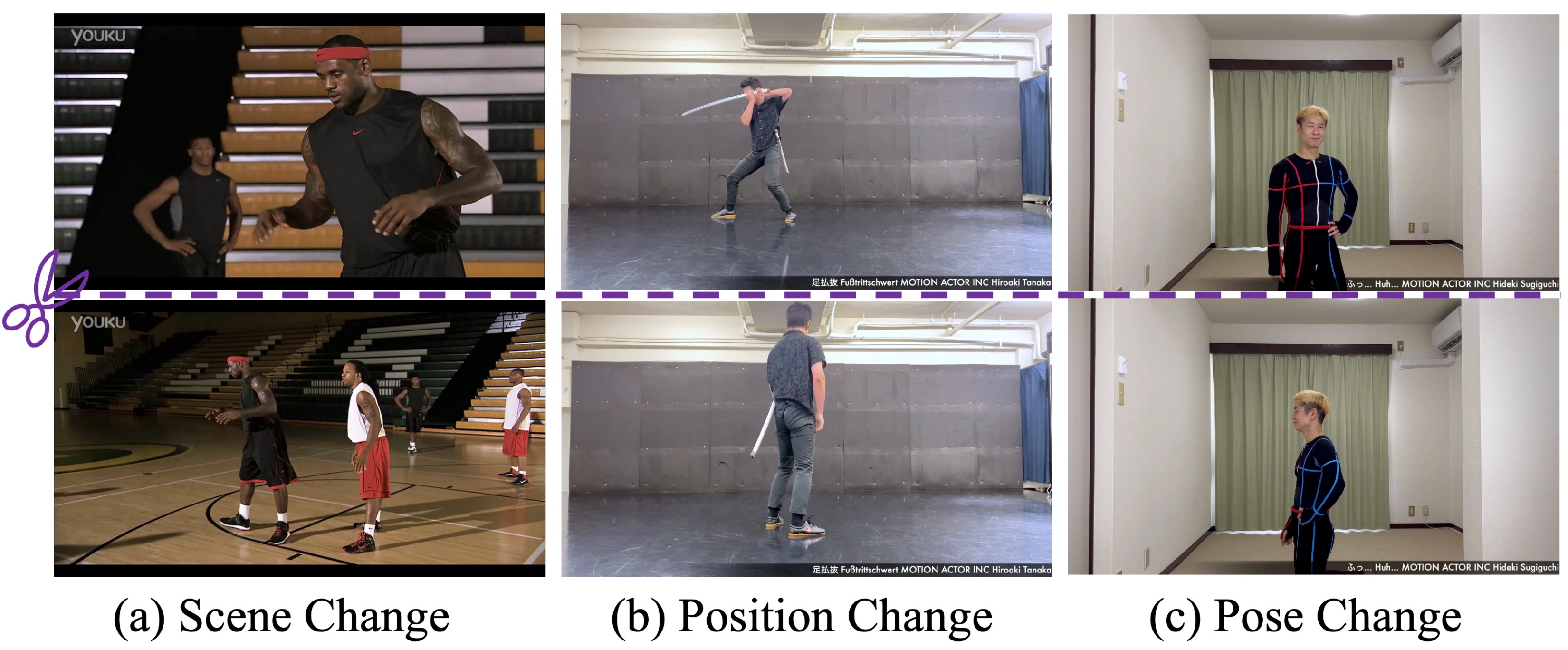}
    \vspace{-0.6em}
    \caption{\textbf{Shot transition detection examples}. Examples (a), (b), and (c) illustrate multi-shot scenarios in online videos. (a) shows scene transitions detectable by SceneDetect. (b) illustrates significant position changes undetectable by SceneDetect but resolvable with bbox tracking-based method. (c) shows pose or orientation transition, requiring pose tracking-based methods as they cannot be addressed by either SceneDetect or bbox tracking.}
    \vspace{-1em}
    \label{fig:shot_transition}
\end{figure}

Our algorithm begins with shot transition detection in one video. As shown in~\cref{fig:pipeline}, the \textit{shot transition detector} has three key components, scene transition detector, bounding box (\aka bbox) tracking, and human keypoints tracking. 
(1) \textit{Scene change transition detector.} Initially, we employ the SceneDetect~\cite{huang2020movienet} algorithm to identify scene changes based on significant variations in the background. However, the SceneDetect fails to detect shot transitions when background changes are unnoticeable, illustrated in \cref{fig:shot_transition}. Subsequently, we leverage the following modules to bridge the gap. (2) \textit{Bbox tracking for shot transition.} As a shot change often accompanies with a sudden change of human subject size, we track humans in a video via mmtracking~\cite{mmtrack2020}. Consequently, we compute the Intersection over Union (IoU) between neighbor bboxes and identify a shot transition when the IoU falls smaller than a manually tuned threshold. 
(3) \textit{Human pose tracking for shot transition detection.} To achieve a finer granularity, we additionally introduce human 2D KPTs to detect extreme corner shot changes in a video. By thresholding the IoU of corresponding keypoints between neighbor frames, we can accurately identify shot transitions even with subtle human movements. 

As each separate module cannot identify all kinds of shot transitions, the three modules are jointly used to clip a video into several sub-sequences serially. 


\subsection{Human Motion and Camera Pose Estimation For Each Shot}

\label{sec:initialization}

After obtaining the clipped videos, our next goal is to estimate the camera pose and SMPL parameters in the world coordinates for each clipped video. 
The estimated camera pose and motions for each shot will be used to construct the whole motion sequence in the next stage (\cref{sec:alignment}). 

\noindent\textbf{How to estimate the camera parameters accurately?} Our approach for camera parameter calculation is based on a visual odometry (VO) estimation method, LEAP-VO~\cite{chen2024leap}. 
Utilizing the CoTracker method~\cite{karaev2023cotracker}, LEAP-VO estimates the visibility and trajectories of $N$ selected points by analyzing image gradients across the video sequence.
LEAP-VO subsequently computes confidence scores for each trajectory, retaining only those with high confidence while discarding trajectories shorter than a predefined threshold. 
The remaining trajectories undergo bundle adjustment (BA) within a fixed window size to estimate the camera poses.

However, simply applying LEAP-VO in the camera estimation process is still unsatisfactory in most human-centric scenarios. The primary limitation stems from the dynamic movements of human subjects, which typically occupy a substantial portion of each image in human-centric videos. This dynamic presence introduces noise into the camera pose estimation in world coordinates, as the estimation process relies heavily on the relationship between the camera and the static environment. To address this issue, we propose a Masked LEAP-VO algorithm. Our approach involves inputting the image $I_t$ and the human bbox at frame $t$ into SAM~\cite{kirillov2023segany} to generate a human mask. We then assign a visibility value of zero to points within the human mask, effectively excluding these trajectories from the BA process. For clarity, we denote $S_{BA}$ as the window size of BA, $\hat{n}$ denotes the number of filtered point trajectories, and $w_{ij, \hat{n}}$ as the normalized weight based on confidence score and visibility. For estimating the camera poses $\mathbf{G}=\{\mathbf{R}, \mathbf{T}\}$ of orientation and translation, the reprojection loss function for BA can then be formulated as follows,
\begin{equation*}
    \scriptsize
    \begin{aligned}
    \scriptsize
    \mathbf{G} = \mathop{\arg\min}\limits_{\mathbf{G},  d_{i, \hat{n}}} \sum_i \sum_{j \in |i - j| \leq S_{BA}} \sum_{\hat{n}} w_{ij, \hat{n}} || \mathcal{F}(\mathbf{G}_i, \mathbf{G}_j, d_{i, \hat{n}}) - \Pi_{ij}(\mathbf{p}_{i, \hat{n}}) ||,
    \end{aligned}
    \vspace{-0.3em}
    \label{eq:leapvo_ba}
\end{equation*}
%
where $\mathcal{F}(\mathbf{G}_i, \mathbf{G}_j, d_{i, \hat{n}})$ denotes the point positions calculated by camera pose $\mathbf{G}$ at frame $i$ and $j$ with depth $d_{i, \hat{n}}$.
$\Pi_{ij}(\mathbf{p}_{i, \hat{n}})$ denotes the position for project position of $\mathbf{p}_{i, \hat{n}}$ from frame $i$ to $j$. Consequently, we obtain the camera rotation $\mathbf{R}_t$ and translation $\mathbf{T}_t$ from camera pose $\mathbf{G}_t$ at $t$.

\noindent\textbf{Recovering human motion in world coordinates with estimated camera parameters.} Given an input video, we feed the estimated camera parameters ($\mathbf{R}_t$ and $\mathbf{T}_t$) into the state-of-the-art motion recovering model, GVHMR~\cite{shen2024gvhmr},
\vspace{-0.5em}
\begin{equation}
    \theta_t^w, \beta_t^w, \Gamma_t^w, \tau_t^w = \mathtt{GVHMR}(I_t, \mathbf{R}_t, \mathbf{T}_t).
\vspace{-0.5em}
\end{equation}
Initialized human parameters $\theta_t^w, \beta_t^w, \Gamma_t^w, \tau_t^w$ and camera parameters $\mathbf{R}_t, \mathbf{T}_t$ will input to human motion alignment.

\subsection{Aligning Human Motion Between Shots}
\label{sec:alignment}

Based on initialized world motion for each individual shot, the subsequent question is \textit{how to merge discontinuous motions from different shots into a continuous motion sequence as a whole in world coordinates}. A straightforward solution is to align all motion sequences to the world coordinate system of the first shot. However, finding the correspondence between different shots is still under-explored and challenging. To resolve this issue, we decompose the motion parameters into camera-dependent and camera-independent ones. The former (\cref{sec:orientation_alignment}) achieves alignment between shots via human orientation alignment based on camera calibration, whereas the latter (\cref{sec:pose_alignment}) is a trainable module to enhance the continuity of human motion sequence. These two key designs ensure a consistent motion sequence between frames when encountering shot transitions.

\subsubsection{Aligning Human Orientations Between Shots}
\label{sec:orientation_alignment}

\definecolor{number1}{HTML}{d8bfd8}
\definecolor{number2}{HTML}{caf0fe}
\definecolor{number3}{HTML}{00c7fc}

\begin{figure}[!t]
    \centering
    \includegraphics[width=0.95\linewidth]{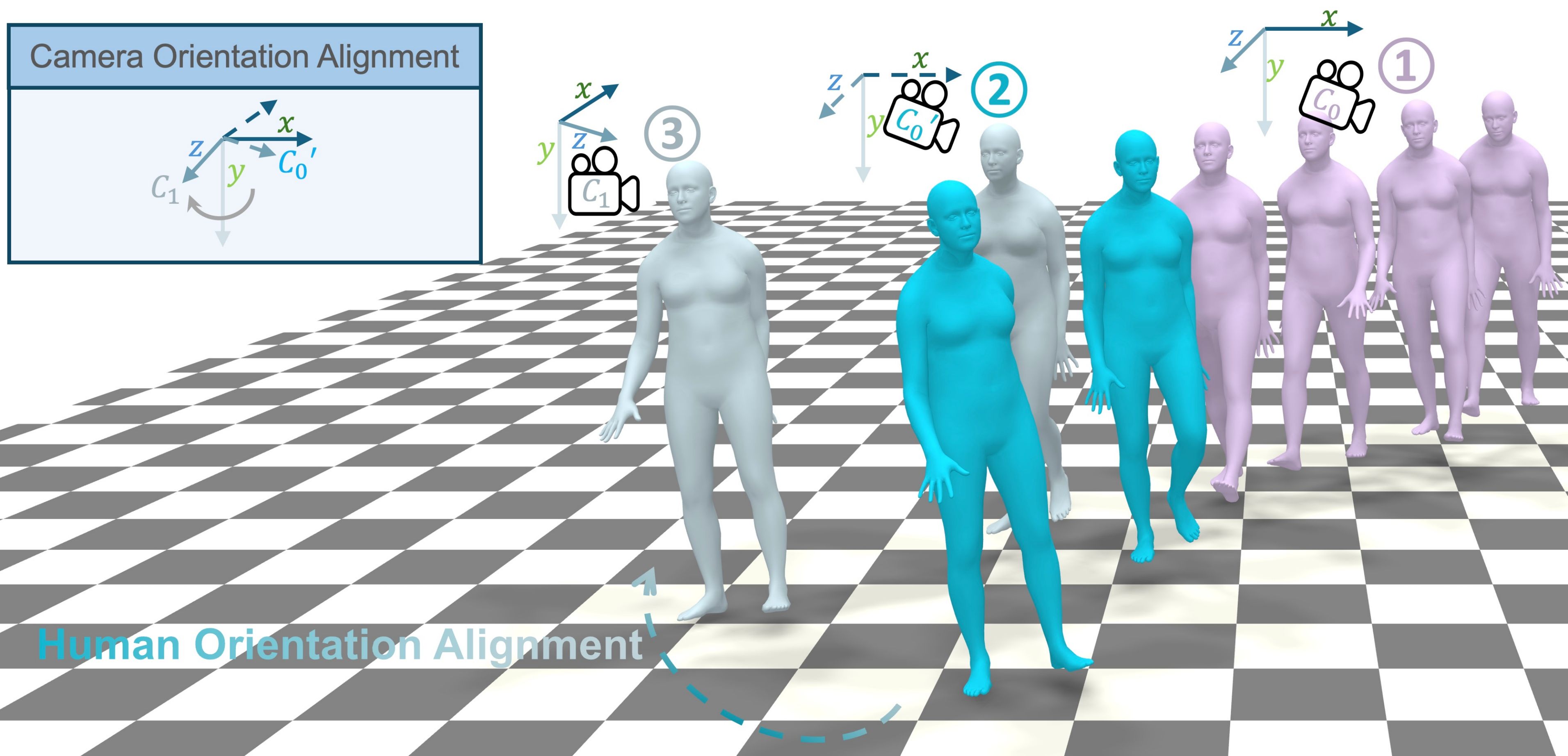}
    \caption{\textbf{Human orientation alignment module}. Following a shot transition after the foremost purple human mesh (shot \textcolor{number1}{\ding{172}} captured by camera $C_0$), the unaligned (blue) and aligned (green) motions are captured as shot \textcolor{number2}{\ding{173}} and shot \textcolor{number3}{``\ding{174}''} by camera $C_0^{'}$ and $C_1$, respectively. $C_0^{'} = C_0$. To achieve human orientation alignment from shot \textcolor{number1}{``\ding{172}''} to \textcolor{number3}{``\ding{174}''}, the camera rotation matrix from $C_0^{'}$ to $C_1$ is computed and applied as the offset of human orientation.}
    \label{fig:orientation}
\end{figure}

After obtaining the initial SMPL and camera parameters $\{\theta_t^i, \beta_t^i, \Gamma_t^i, \tau_t^i, \mathbf{R}_t^i, \mathbf{T}_t^i\}$ for each shot, directly concatenating motions between shots result abrupt changes of human poses and orientations. 
To address this issue, we introduce the \textit{Orientation Alignment Module} (OAM), as shown in \cref{fig:orientation}, to align human orientations. As the whole motion sequence is continuous, we have the following assumption.
\vspace{-1.5em}
\begin{assumption}
\label{assum1}
    Human orientations and translations during the shot transition in world coordinates are continuous.
\end{assumption}
\vspace{-0.2em}
To align the orientations between two frames with shot transition under Assumption~\ref{assum1}, we decompose the human orientation with shot transitions in world coordinates as,
\vspace{-0.5em}
\begin{equation}
    \mathtt{R}(\Gamma_{\text{world}}) = \mathbf{R}_{\delta_{\text{cam}}} \mathtt{R}(\Gamma_{\text{view}}),
\vspace{-0.5em}
    \label{eq:rotation}
\end{equation}
where $\mathbf{R}_{\delta_{\text{cam}}}$ represents the camera rotation on the Y-axis between current $t$-th and previous $t-1$-th frame, $\Gamma_{\text{view}}$ denotes the human orientation estimated by the current shot, and $\mathtt{R}(\cdot): \mathbb{R}^{3}\rightarrow \mathbb{R}^{9}$ is the mapping from axis angle to rotation matrix. As $\Gamma_{\text{view}}$ in current shot can be estimated independently, mentioned in~\cref{sec:initialization}, obtaining accurate $\Gamma_{\text{world}}$ in~\cref{eq:rotation} remains a key challenge to estimate the relative camera rotation $\mathbf{R}_{\delta_{\text{cam}}}$ between frames in shot transitions. 

\noindent\textbf{Estimating the relative camera pose $\mathbf{R}_{\delta_{\text{cam}}}$ between transition frames.} Different from our approach of estimating camera pose in each shot (\cref{sec:initialization}), we do not mask the human subject when estimating camera rotation $\mathbf{R}_{\delta_{\text{cam}}}$. Instead, we use human 2D KPTs as explicit feature matching. Specifically, we filter out unmatched keypoints based on their visibility and unaligned direction using RANSAC~\citep{fischler1981ransac}, effectively addressing camera pose estimation during shot transitions. This procedure is referred to as \textit{Camera Calibration} (\aka epipolar-geometry-based camera extrinsics estimation), and is detailed below.

In \textit{Camera Calibration}, we assume that the human translations remain unchanged across the shot transition, implying that only the camera's orientation changes (\ie Assumption~\ref{assum1}). 
Consequently, we calculate the orientation offset by determining the change in camera orientation using camera calibration.
We begin by extracting human 2D KPTs from two consecutive frames during the shot transition. 
Due to the shot transition, the visibility of 2D KPTs may vary, \eg occlusion in some shots. Therefore, we employ EDPose~\cite{yang2023explicit} to filter out invisible 2D KPTs between shot transition frames. 
Subsequently, RANSAC identifies matching 2D KPTs corresponding to the most possible camera rotation direction.
These matched 2D KPTs facilitate the estimation of the aligned camera rotation $\mathbf{R}_{\delta_{\text{cam}}}$. The detailed estimation process is as follows.

We denote the detected 2D KPTs of two frames in the shot transition as $\mathbf{S}_1 = [(x_1^{(1)}, y_1^{(1)}), (x_1^{(2)}, y_1^{(2)}), \cdots, (x_1^{(N)}, y_1^{(N)})]^{\top} \in \mathtt{R}^{2\times N}$ and  $\mathbf{S}_2 = [(x_2^{(1)}, y_2^{(1)}), (x_2^{(2)}, y_2^{(2)}), \cdots, (x_2^{(N)}, y_2^{(N)})]^{\top} \in \mathtt{R}^{2\times N}$. The essential matrix $\mathbf{E} = [\mathbf{T}]_{\times}\mathbf{R}$ should satisfy the following orthogonal property such that, 
\vspace{-0.1em}
\begin{equation}
\mathbf{S}_1^{\top}\mathbf{E}\mathbf{S}_2 = \mathbf{0}.
    \label{eq:e_matrix}
\end{equation}
Once $\mathbf{E}$ is obtained by solving~\cref{eq:e_matrix}, we enforce the rank-2 constraint on $\mathbf{E}$ through SVD decomposition and subsequently derive the aligned camera rotation $\mathbf{R}_{\delta_{\text{cam}}}$ between two frames (\cf Hartley \etal~\cite{hartley2003triangulation} for more details).  

In summary, we reformulate the alignment problem of human orientation in shot transitions as estimating the relative camera rotation $\mathbf{R}_{\delta_{\text{cam}}}$ between frames. Accordingly, we obtain the camera rotation $\mathbf{R}_{\delta_{\text{cam}}}$ via camera calibration.


\subsubsection{Aligning Human Poses Between Shots}
\label{sec:pose_alignment}


\begin{figure}[!t]
    \centering
    \begin{overpic}[width=0.98\linewidth]{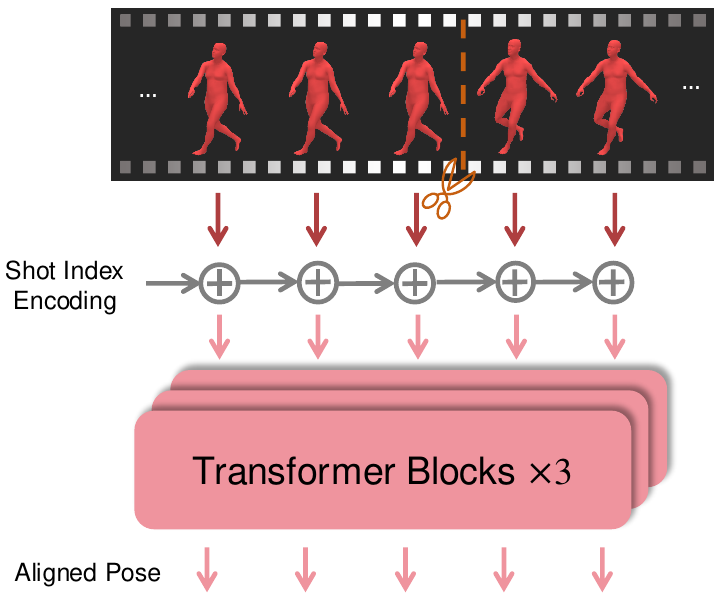}
        \put(1, 70){$\{\theta_t\}_{t=1}^{T}$}
        \put(90, 2){$\{\phi_t\}_{t=1}^{T}$}
    \end{overpic}
    \vspace{-0.6em}
    \caption{\textbf{\mshmr Structure.} The initial human pose parameters $\theta$ across multiple video shots are input into a transformer with shot-index-based positional encoding. This enables \mshmr to generate consistent human poses across all shots in the video.}
    \vspace{-1em}
    \label{fig:mshmr}
\end{figure}

In shot transition, video sequences recorded by two shots are often with various occlusions. However, unoccluded body parts in two shots can be complementary to each other for motion alignment. Thus, we introduce the \textit{multi-shot HMR} (\textit{ms}-HMR, \ie $\mathtt{E}_M(\cdot)$) module to refine the whole motion sequence. As shown in~\cref{fig:mshmr}, the \textit{ms}-HMR is a Transformer encoder-like architecture, whose input and output are the estimated global motion and the refined global motion, respectively. The process can be formulated as, 
\begin{equation}
    \phi_1, \phi_2, \cdots, \phi_T = \mathtt{E}_M(\theta_1, \theta_2, \cdots, \theta_T), 
\end{equation}
where $\phi_{*}$ denotes the refined motion of each frame. With this design, our method can adapt to diverse occlusions of human body brought by shot transitions.

\subsection{Post-processing Module for Motion Integration}

\label{sec:network}

\noindent\textbf{Trajectory Predictor and Trajectory Refiner.} Based on the aligned human pose and orientation, we introduce a bi-directional LSTM trajectory predictor to recover foot-ground contact probabilities $p_t^c$, and root velocity $v_t$ as,
\vspace{-0.2em}
\begin{equation}
\begin{aligned}
\small
     p_t^c, v_t = \mathtt{LSTM}(\phi_1^m, \Gamma_1, \mathtt{F}({I_1}), \phi_2^m, \Gamma_2, \mathtt{F}({I_2}), \cdots,\\
     \phi_T^m, \Gamma_T, \mathtt{F}({I_T})),
\end{aligned}
\end{equation}
where $\mathtt{F}({\cdot})$ denotes the image feature of each frame extracted by ViT~\cite{dosovitskiy2020vit}. Accordingly, the contact probabilities $p_t^c$, and velocity $v_t$ are supervised by the ground-truth labels with MSE loss and are used to reconstruct the trajectory. Besides, we extend the trajectory refiner in WHAM~\cite{shin2024wham} to alleviate foot sliding problem in our estimated trajectory.

\begin{table*}[!t]
\scriptsize
\centering
\setlength{\tabcolsep}{2pt}
\renewcommand{\arraystretch}{1}
\resizebox{\linewidth}{!}{
\begin{tabular}{llcccccccccccccccccc}
\toprule
\multirow{2}{*}{\centering Dataset} & \multirow{2}{*}{\centering Models} & \multicolumn{6}{c}{2-Shot} & \multicolumn{6}{c}{3-Shot} & \multicolumn{6}{c}{4-Shot} \\ 
\cmidrule(lr){3-8} \cmidrule(lr){9-14} \cmidrule(lr){15-20}
 & & PA.$\downarrow$ & WA.$\downarrow$ & RTE$\downarrow$ & ROE$\downarrow$ & Jitter$\downarrow$ & F.S.$\downarrow$ & PA.$\downarrow$ & WA.$\downarrow$ & RTE$\downarrow$ & ROE$\downarrow$ & Jitter$\downarrow$ & F.S.$\downarrow$ & PA$\downarrow$ & WA$\downarrow$ & RTE$\downarrow$ & ROE$\downarrow$ & Jitter$\downarrow$ & F.S.$\downarrow$ \\ 
\midrule
\multirow{2}{*}{\msaist} 

& SLAHMR~\citeyearpar{ye2023decoupling} & 72.34 & 341.75 & 9.62 & 96.26 & 62.59 & 3.26 & 80.35 & 510.77 & 10.33 &  101.36 & 72.39 & 4.43 & 90.32 & 803.69 &12.11 & 104.07 & 80.37 & 16.52  \\

& WHAM~\citeyearpar{shin2024wham} & 65.34 & 336.82 & 4.39 & 84.48 & \textbf{25.24} & 2.75 & 78.68 & 451.32 & 5.14 & 89.84 & \textbf{24.06} & \textbf{2.99} & 102.88 & 603.93 & 5.57 & 90.07 & \textbf{26.29} & \textbf{3.62}  \\

& GVHMR~\citeyearpar{shen2024gvhmr} & 60.72 & 231.36 & 6.20 & 96.58 & 34.87 & 7.65 & 70.33 & 357.16 & 7.55 & 99.69 & 34.46 & 9.42 & 83.77 & 563.17 & 8.96 & 104.53 & 35.67 & 9.78 \\ 
  
 &  \cellcolor{lightgray} \textbf{Ours} & \cellcolor{lightgray} \textbf{36.82} & \cellcolor{lightgray} \textbf{121.35} & \cellcolor{lightgray} \textbf{2.56} &  \cellcolor{lightgray} \textbf{69.23} &  \cellcolor{lightgray} 33.27 &  \cellcolor{lightgray} \textbf{2.66} &  \cellcolor{lightgray} \cellcolor{lightgray} \textbf{38.52} & \cellcolor{lightgray} \textbf{141.38} & \cellcolor{lightgray} \textbf{3.64} &  \cellcolor{lightgray} \textbf{67.71} & \cellcolor{lightgray} 35.07 &  \cellcolor{lightgray} 3.55 & \cellcolor{lightgray} \textbf{39.63} & \cellcolor{lightgray} \textbf{161.52} & \cellcolor{lightgray} \textbf{4.55} &  \cellcolor{lightgray} \textbf{70.31} &  \cellcolor{lightgray} 39.49 &  \cellcolor{lightgray} 4.09  \\

 \specialrule{0em}{0.0pt}{0.0pt}  
\midrule
\multirow{2}{*}{\msm}

 & SLAHMR~\citeyearpar{ye2023decoupling} & 80.67 & 352.61 & 16.67 & 111.97 & 37.80 & 7.93 & 97.15 & 562.10 & 16.91 & 118.46 & 52.23 & 9.96 & 107.90 & 748.58 & 17.85 & 116.72 & 65.15 &  11.58  \\

  & WHAM~\citeyearpar{shin2024wham} & 71.32 & 313.58 & 11.41 & 82.42 & \textbf{18.40} & 5.09 & 79.51 & 423.98 & 12.36 & 84.85 & 18.87 & 5.03  & 90.50 & 512.66 & 12.91 & 90.34 & \textbf{18.40} & 5.69  \\

  & GVHMR~\citeyearpar{shen2024gvhmr} & 64.63 & 254.30 & 6.94 & 81.93 & 18.45 & 8.80 & 80.79 & 296.74 & 85.25 & 58.26 & 18.36 & 10.62 & 85.19 & 471.53 & 9.12 & 91.63 & 19.47 & 10.65 \\

   &\cellcolor{lightgray} \textbf{Ours} & \cellcolor{lightgray} \textbf{40.52} & \cellcolor{lightgray} \textbf{132.13} & \cellcolor{lightgray} \textbf{3.65} & \cellcolor{lightgray} \textbf{53.39} & \cellcolor{lightgray}  19.05 & \cellcolor{lightgray} \textbf{4.17} & \cellcolor{lightgray} \textbf{45.35} & \cellcolor{lightgray} \textbf{145.36} & \cellcolor{lightgray} \textbf{5.33} & \cellcolor{lightgray} \textbf{58.26} & \cellcolor{lightgray} \textbf{17.35} & \cellcolor{lightgray}  \textbf{4.62} &
   \cellcolor{lightgray} \textbf{50.59} & \cellcolor{lightgray} \textbf{147.62} & \cellcolor{lightgray}  \textbf{6.20} & \cellcolor{lightgray} \textbf{61.22} &\cellcolor{lightgray}  19.77  & \cellcolor{lightgray} \textbf{5.12}\\
\specialrule{0em}{0.0pt}{0.0pt}  \bottomrule
\end{tabular}
}
\vspace{-0.6em}
\caption{\textbf{Quantitative comparison of different HMR methods on \dataname dataset.} We record the results for \msaist and \msm separately. PA. and WA. means PA-MPJPE and WA-MPJPE respectively, while F.S. is Foot Sliding. Our proposed method has achieved the best performance in PA-MPJPE, WA-MPJPE, RTE and ROE across \dataname among these methods.}
\label{tab:ms_comparison}
\end{table*}

\section{Benchmarking Multi-shot Motion Recovery}
\label{sec:dataset}

\noindent\textbf{Dataset Construction.} To create a multi-shot 3D human motion dataset, we introduce \dataname by processing existing public 3D human datasets with multiple camera settings and ground truth human and camera parameters, specifically AIST~\cite{li2021aistpp} and Human3.6M (H3.6M)~\citep{ionescu2014human36m}.
In our construction pipeline, we randomly separate each original one-shot video into several clips.
Then, we choose each clip from different shots and concatenate them together as one video recorded by multiple shots.
For example, AIST provides each video with eight cameras 
\texttt{C0}, \texttt{C1}, ..., \texttt{C7} from different view point and we choose a video and split it into 5 clips at \texttt{t0}, \texttt{t1}, ..., \texttt{t4}.
For frames in these separated clips, we choose frames shot by a random camera for each clip and combine five clips as one multi-shot video.
Therefore, we construct a multi-shot version of AIST and H3.6M, which are named \msaist and \msm subsets.
%
Then we combine them and name this new dataset \dataname. The detailed statistics of \dataname are shown in \cref{tab:data_stats}. We do not compare with other existing 3D human datasets as they contain limited number of multi-shot videos.
\begin{table}[!t]
    \centering
    \setlength{\tabcolsep}{3pt}
    \resizebox{\linewidth}{!}{%
        \begin{tabular}{lcccccc}
            \toprule
            Dataset & Duration(s) & Videos & FPS & Max Length & Min Length & Shots \\ 
            \midrule
            \dataname & 23.7 & 600 & 30 & 1478 & 314 & 2, 3, 4 \\ 
            \bottomrule
        \end{tabular}
    }
    \vspace{-0.6em}
    \caption{Statistics of the \dataname dataset. By shots, we mean the number of shot transitions in a single video.}
    \vspace{-1.3em}
    \label{tab:data_stats}
\end{table}

\noindent\textbf{Benchmark Evaluation Protocol.} To evaluate the performance of our proposed methods on multi-shot videos, our target is to evaluate metrics for accurately reflecting the performance on videos with shot transitions. 
To this end, we use Root Orientation Error (\aka ROE in $deg$ ) to measure the performance of the proposed method on human orientation alignment across different shots. 
Besides, we use Root Translation Error (\aka RTE in $m$) to assess the performance of the proposed method on global trajectory recovery.
Jitter ($\frac{10m}{fps^3}$) is also used to evaluate the stability of recovered human pose from multi-shot videos.
We also include foot sliding ($cm$), the averaged displacement of foot vertices during contact with the ground, to assess the precision of recovered motion in the world coordinates~\cite{shin2024wham}.

\section{Experiment}
\subsection{Datasets and Metrics} 

\noindent\textbf{Evaluation Datasets.} To evaluate the performance of our proposed pipeline for multi-shot videos, we use \dataname dataset and EMDB-1 dataset~\cite{kaufmann2023emdb} with self-added noise for the evaluation of ablation study. For camera trajectory estimation, we use EMDB-1 and EMDB-2 split~\cite{kaufmann2023emdb} as they contain the GT moving camera trajectory. Our self-created dataset contains 600 multi-shot videos, 42.7K frames, totaling 237 minutes. EMDB-1 split contains 17 video sequences totaling 13.5 minutes and EMDB-2 split contains 25 sequences in a total of 24.0 minutes.

\noindent\textbf{Evaluation Metrics.} For shot detection we use \textit{Recall}, \textit{Precision} and \textit{F1 Score} as evaluation metrics. 
%
For 3D human pose estimation-related tasks, we use ROE, RTE, jitter, and foot-sliding for evaluating the human motion recovery results on multi-shot videos. For the ablation study of our proposed pipeline, we evaluate the Procrustes-aligned Mean Per Joint Position Error (\aka PA-MPJPE) and Per Vertex Error (\aka PVE) as additional metrics besides previous mentioned ones. For camera pose estimation, we use absolute trajectory error (\aka ATE) ($m$), Relative Pose Error (\aka RPE) rotation ($deg$), and RPE translation ($m$).

\subsection{Implementation Details}

The \textit{ms}-HMR, the trajectory, and foot sliding refiner are trained on the AMASS~\cite{mahmood2019amass}, 3DPW~\cite{von20183dpw}, Human3.6M~\cite{ionescu2014human36m}, and BEDLAM~\cite{black2023bedlam} datasets, and evaluated on EMDB and our \dataname. During training, we introduce random rotational noise (ranging from 0 to 1 radian) along the y-axis to the root pose $\Gamma$ and random noise to the body pose $\theta$ at random positions to simulate the inaccuracies of pre-estimated human motions caused by shot transitions in multi-shot videos. This strategy enables the network to robustly recover smooth and consistent human motion from noisy initial parameters.
The benchmark test results were obtained after training for 80 epochs on one NVIDIA-A100 GPU.

\begin{table*}[!t]
    \centering
    \resizebox{0.99\textwidth}{!}{
    \begin{tabular}{l|cccccccccc}
    \toprule
     Methods & PA-MPJPE$\downarrow$ & MPJPE$\downarrow$  & WA-MPJPE$\downarrow$ & W-MPJPE$\downarrow$  &  PVE$\downarrow$& ACCEL$\downarrow$ & RTE$\downarrow$ & ROE$\downarrow$ &Foot Sliding$\downarrow$\\
    \midrule
    Baseline (Concat)      & 106.48 & 141.67 & 273.15 & 553.67 & 122.15 & 6.14 & 10.86 & 91.55 & 14.91\\
    w/o \mshmr          & 78.24 & 101.52 & 246.42 & 436.57 & 85.77  & 5.87 &3.89 & 50.63 & 3.54\\
    w/o OAM               & 73.56 & 92.13 & 243.65 & 425.18 & 79.64  & 5.67 &6.61 & 76.74 & 4.45 \\
    w/o trajectory predictor & 50.49 & 83.68 & 231.75 & 432.17 & 75.77 & 5.75 & 5.52 & 47.68 & 4.96 \\
    w/o trajectory refiner  & 50.49 & 83.68 & 198.58 & 397.65 & 75.77  & 5.23 &4.06 & 47.68 & 7.84\\
    \rowcolor{lightgray} \textbf{\methodname (Ours)} &50.49 & 83.68 & 194.77 & 393.21 & 75.77 & 5.16 & 3.54 & 47.68 & 3.28\\
    \specialrule{0em}{0.0pt}{0.0pt}  \bottomrule
    \end{tabular}
    }
    \vspace{-0.6em}
    \captionof{table}{\textbf{Ablation studies on different combinations of \methodname modules.} We evaluate the contributions of each key module in \methodname (\mshmr, OAM, trajectory predictor and trajectory refiner) to overall performance using various human motion recovery metrics. The complete \methodname model achieves state-of-the-art performance with minimal foot sliding and robust global motion recovery.}
    \label{tab:ablation_results}
\end{table*}

\begin{table}[!h]
    \centering
    \resizebox{\linewidth}{!}{
    \begin{tabular}{l|ccc}
    \toprule
    \multirow{2}{*}{Methods} & \multicolumn{3}{c}{\dataname} \\ 
    \cmidrule(lr){2-4}
     & Recall$\uparrow$ & Precision$\uparrow$  & F1 Score$\uparrow$ \\
    \midrule
    Scenes Detect (SD)~\cite{huang2020movienet} & 0.74 & 0.72 & 0.70  \\
    SD+Bbox Tracking (Bbox) & 0.88 & 0.85 & 0.86 \\
    SD+Bbox+Pose Tracking & \textbf{0.96} & \textbf{0.88} & \textbf{0.92} \\
    \bottomrule
    \end{tabular}
    }
    \vspace{-0.6em}
    \caption{\textbf{Comparison between difference shot detection algorithms.} We evaluate our shot transition detector on our proposed multi-shot video human motion dataset \dataname. 
    \vspace{-1em}
    \label{tab:shot_det_results}}
\end{table}

\subsection{Main Results: Comparison of Global Human Motion Recovery Results on the Benchmark}
We compare our proposed method \methodname with several state-of-the-art HMR methods (SLAHMR~\cite{ye2023decoupling}, WHAM~\cite{shin2024wham} and GVHMR~\cite{shen2024gvhmr}) on our proposed benchmark \dataname. As illustrated in~\cref{tab:ms_comparison}, our proposed method has achieved the best performance for PA\&WA-MPJPE, RTE and ROE through videos with all numbers of shots across \msaist and \msm, indicating that our method reconstructs both the global human motion and orientations in the world coordinates more accurately and robustly. For the foot sliding metric, our method also performs as the best on \msm across all numbers of shots.


\begin{table}[!t]
    \centering
    \resizebox{\linewidth}{!}{
    \begin{tabular}{l|ccc}
    \toprule
    Methods & ATE$\downarrow$ & RPE Trans.$\downarrow$  & RPE Rot.$\downarrow$ \\
    \midrule
    DPVO (w/o mask) & 0.48 & 1.85 & 1.06  \\
    Masked DPVO & \textbf{0.48} & 1.57 & 0.97 \\
    \rowcolor{lightgray} LEAP-VO (w/o mask) & 0.50 & 0.93 & 0.97 \\
    \rowcolor{lightgray} \textbf{Masked LEAP-VO} & 0.51 & \textbf{0.92} & \textbf{0.95} \\
    \specialrule{0em}{0.0pt}{0.0pt}  \bottomrule
    \end{tabular}
    }
    \vspace{-0.6em}
    \caption{\textbf{Camera tracking results on EMDB-1~\cite{kaufmann2023emdb}.} Our method has achieved $\sim 50\% \downarrow$ on RPE trans. than that of the original DPVO and achieve the best performance in RPE rot. metrics. 
    \vspace{-1em}
    \label{tab:cam_emdb1}}
\end{table}

\begin{figure*}[!t]
    \centering
    \includegraphics[width=0.99\linewidth]{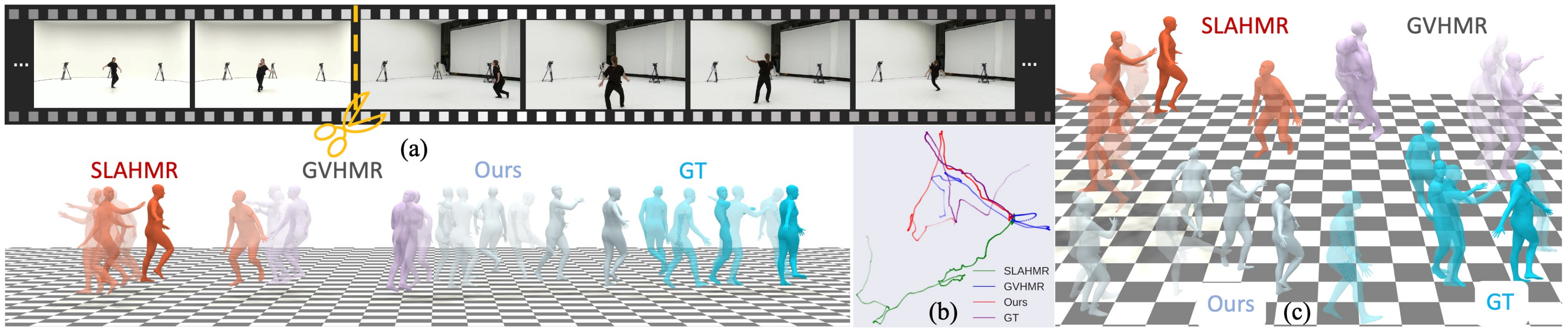}
    \vspace{-0.6em}
    \caption{\textbf{Qualitative comparison of different HMR methods on \dataname dataset.} The side view of the rendered mesh for input mutli-shot video is shown in (a), while the top view is shown in (c). We also draw the comparison of the human trajectory as shown in (b). Our method is the most similar as GT in both rendered motion and trajectories among these methods.}
    \label{fig:vis_comparison}
\end{figure*}

\subsection{Ablation Studies}
\label{sec:ablation_study}
\noindent \textbf{Human-centric Scene Shot Boundary Detection Evaluation.} To evaluate the performance of our proposed \textit{Shot Transition Detector}, we test the algorithm on our proposed multi-shot human motion recovery benchmark and compare the output frame list of shot transitions with the ground truth (GT) of our dataset. As shown in~\cref{tab:shot_det_results}, by applying the proposed finer granularity shot detection methods, the number of recall, precision, and F1 score all increases consistently. The combination of three steps (ScenesDetect, bbox tracking, and pose tracking) has achieved 0.96, 0.88, and 0.92 on the recall, precision, and F1 score, respectively, which indicates a comparable performance in shot boundary detection. Besides, as can be seen in the results, the latter two steps of shot detection contribute to the fine-grained final results significantly and jointly.

\noindent \textbf{Key modules in the Proposed Method.} We compare our methods with four variants on EMDB with noise dataset, as shown in~\cref{tab:ablation_results}, \textit{ms}-HMR is the key component for the improvement in PA-MPJPE and PVE, which indicates a more accurate modeling of the whole motion sequence. This design serves as a recovery module to estimate some invisible body parts in some shots. Additionally, the orientation alignment module (\textit{OAM}, in~\cref{sec:alignment}) is also a critical block for accurate human orientation estimation, indicated by the metric ROE. This module helps to model the global human motion between shots. For foot sliding, the results in~\cref{tab:ablation_results} also show that the trajectory refiner (\cref{sec:network}) in our method helps mitigate the foot sliding issue.

\begin{table}[!t]
    \centering
    \resizebox{\linewidth}{!}{
    \begin{tabular}{l|ccc}
    \toprule
     Methods & ATE$\downarrow$ & RPE Trans.$\downarrow$  & RPE Rot.$\downarrow$ \\
    \midrule
    DPVO (w/o mask) & \textbf{0.48} & 1.07 & 1.26  \\
    Masked DPVO & 0.50 & 0.86 & 1.21 \\
    \rowcolor{lightgray} LEAP-VO (w/o mask) & 0.50 & 0.83 & 1.21 \\
    \rowcolor{lightgray}\textbf{Masked LEAP-VO} & 0.49 & \textbf{0.83} & \textbf{1.19} \\
    \specialrule{0em}{0.0pt}{0.0pt}  \bottomrule
    \end{tabular}
    }
    \vspace{-0.6em}
    \caption{\textbf{Camera tracking results on EMDB-2~\cite{kaufmann2023emdb}.} Our method maintains its comparable performance in both RPE Trans. and RPE Rot. metrics, illustrating its effectiveness.
    \vspace{-1.2em}
    \label{tab:cam_emdb2}}
\end{table}

\begin{table}[!t]
    \centering
    \resizebox{\linewidth}{!}{
    \begin{tabular}{l|ccccc}
    \toprule
     Methods & WA-MPJPE$\downarrow$ & W-MPJPE$\downarrow$ & RTE$\downarrow$ & Jitter$\downarrow$ & F.S.$\downarrow$\\
    \midrule
    DPVO      & 305.40 & 117.10 & 5.10 & 17.90 & 3.50\\
    Masked DPVO          & 303.90  & 116.40  & 4.10 & 17.40 & 3.50\\
    \rowcolor{lightgray} LEAP-VO               & 284.10  & 112.80  & 3.10 & 16.30 & 3.50 \\
    \rowcolor{lightgray} \textbf{Masked LEAP-VO} & \textbf{283.70}  & \textbf{112.70}  & \textbf{3.10} & \textbf{16.30} & \textbf{3.50}\\
    \specialrule{0em}{0.0pt}{0.0pt}  \bottomrule
    \end{tabular}
    }
    \vspace{-0.6em}
    \caption{\textbf{Global motion recovery results on EMDB-2~\cite{kaufmann2023emdb}.} We input estimated camera parameters from different methods into GVHMR for the comparison on human motion recovery metrics. 
    \vspace{-1em}
    \label{tab:cam_human}}
\end{table}

\begin{table*}[!t]
    \centering
    \setlength{\tabcolsep}{4pt}
    \renewcommand{\arraystretch}{1}
    \resizebox{\linewidth}{!}{
    \begin{tabular}{lcccccccccccccccccc}
    \toprule
    \multirow{2}{*}{Methods} & \multicolumn{6}{c}{2-shot} & \multicolumn{6}{c}{3-shot} & \multicolumn{6}{c}{4-shot} \\ 
    \cmidrule(lr){2-7} \cmidrule(lr){8-13} \cmidrule(lr){14-19}
    & PA.$\downarrow$ & WA.$\downarrow$ & RTE$\downarrow$ & ROE$\downarrow$ & Jitter$\downarrow$ & F.S.$\downarrow$ & PA.$\downarrow$ & WA.$\downarrow$ & RTE$\downarrow$ & ROE$\downarrow$ 
    & Jitter$\downarrow$ & F.S.$\downarrow$ & PA.$\downarrow$ & WA.$\downarrow$ & RTE$\downarrow$ & ROE$\downarrow$ & Jitter$\downarrow$ & F.S.$\downarrow$\\ 
    \midrule
    
    SLAHMR~\citeyearpar{ye2023decoupling}  & 75.54 & 326.15 & 9.35 & 94.63 & 63.74 & 3.15 & 81.22 & 523.98 & 11.59 & 105.44 & 74.12 & 4.55 & 93.45 & 811.31 & 14.85 & 109.56 & 80.69 & 14.12\\
    
    WHAM~\citeyearpar{shin2024wham} & 65.14 & 339.85 & 4.65 & 81.12 & \textbf{25.29} & 2.82 & 83.10 & 501.29 & 7.82 & 89.88 & \textbf{25.85} & 3.05 & 100.20 & 801.63 & 5.97 & 94.36 & \textbf{27.29} & 3.98 \\
    
    GVHMR~\citeyearpar{shen2024gvhmr} & 66.18 & 234.96 & 5.82 & 92.13 & 35.77 & 4.72 & 72.63 & 359.99 & 6.42 & 97.45 & 35.63 & 5.85 & 90.52 & 588.40 & 9.02 & 96.75 & 31.66 & 9.52 \\ 
    
    \rowcolor{lightgray} 
    \textbf{Ours} & \textbf{33.96} & \textbf{109.72} &\textbf{2.12} & \textbf{66.58} & 33.49 & \textbf{1.69} & \textbf{40.12} & \textbf{131.39} & \textbf{3.94} & \textbf{68.33} & 31.82 & \textbf{3.01} & \textbf{42.64} & \textbf{165.23} & \textbf{4.32} & \textbf{69.15} & 32.22 & \textbf{3.52}\\
    
    \bottomrule
    \end{tabular}
    }
    \vspace{-0.6em}
    \caption{\textbf{Quantitative results on \dataname (20)}. We select 20 videos from \dataname and generate 2-, 3-, and 4-shot versions for each, yielding a total of 60 videos for evaluation. PA. and WA. means PA-MPJPE and WA-MPJPE respectively, while F.S. is Foot Sliding. The results follow a similar trend to those on original \dataname, showcasing \dataname's effectiveness in evaluating different HMR methods.}
    \label{tab:ms_comparison_60}
    \vspace{-0.1in}
\end{table*}

\noindent \textbf{Comparison on Camera Trajectory Estimation.}
To evaluate the performance of our proposed camera trajectory estimation method \textbf{Masked LEAP-VO}, we evaluate the camera trajectory accuracy on EMDB-1 and EMDB-2. 
For more convenient comparison, we introduce two baselines, DPVO~\cite{teed2023deep}, which has been widely used in HMR methods such as WHAM~\cite{shin2024wham} and GVHMR~\cite{shen2024gvhmr}, and LEAP-VO~\cite{chen2024leap}.
To provide more intuition about the insights of masking dynamic humans in the video, we also implement a variant, Masked DPVO, by applying SAM at the patchify stage of DPVO to exclude patches containing human pixels.
As shown in~\cref{tab:cam_emdb1} and~\cref{tab:cam_emdb2}, compared with baseline methods, our key design of masking dynamic human subjects improves the result in both RPE Translation and RPE Rotation while maintaining competitive ATE. This result indicates the effectiveness of the design of masking dynamic human subjects in the process of camera trajectory estimation. Compared with the DPVO baseline, our method achieves $\sim50\%\downarrow$ RPE translation on EMDB-1. In addition, we run GVHMR~\cite{shen2024gvhmr} on EMDB-2 with different estimated camera trajectories. The results is shown in \cref{tab:cam_human}, which further illustrates the effectiveness of our method.

\noindent \textbf{Further evaluation on \dataname.} 
To ensure a balanced evaluation and evaluate potential data bias in the \dataname dataset, where videos in the 2-, 3-, and 4-shot categories are mutually exclusive, we create multiple versions of the same video across different shot categories for further evaluation. Specifically, we randomly select 20 videos and create their 2-, 3-, and 4-shot versions, resulting in a total of 60 videos. We then evaluate SLAHMR~\cite{ye2023decoupling}, WHAM~\cite{shin2024wham}, GVHMR~\cite{shen2024gvhmr}, and our method on these videos. The results, presented in \cref{tab:ms_comparison_60}, follow the same evaluation pattern as those on the full \dataname dataset in \cref{tab:ms_comparison}, which indicates the effectiveness of our proposed dataset in evaluating different human motion recovery methods.

\section{Conclusion and Discussion}

\label{sec:conclusion}
\noindent \textbf{Conclusion.} In this paper, we introduce \methodname, the first framework designed for human motion recovery from multi-shot videos in world coordinates. \methodname addresses the challenges inherent in multi-shot videos by incorporating three key components: an enhanced camera trajectory estimation method called masked LEAP-VO, a human motion alignment module that ensures consistency across different shots, and a post-processing module for seamless motion integration. Extensive experiments demonstrate that \methodname outperforms existing human motion recovery methods across various benchmarks, achieving state-of-the-art accuracy on our newly created multi-shot human motion dataset, \dataname.

\noindent \textbf{Limitations and Future Work.} While \methodname represents an dvancement in human motion recovery from multi-shot videos in world coordinates, its performance may decline when faced with an excessive number of shot transitions. Despite these challenges, \methodname provides a solid baseline for human motion recovery from multi-shot videos and can be employed in annotating \textit{markerless} human motion datasets. Our newly introduced dataset, \dataname, offers a valuable benchmark for evaluating general human motion recovery methods in world coordinates, especially regarding their performance on multi-shot videos. Based on the proposed method, our future work aims to enlarge the related datasets for larger-scale motion databases. 

\section*{Acknowledgement} 

This work was partially funded by the Shenzhen Science and Technology Project under Grant KJZD20240903103210014. The author team would also like to convey sincere thanks to Ms. Yaxin Chen from IDEA Research for the expressive dance motion used in the demo presentation. 
\clearpage
\clearpage
\clearpage
{
    \small
    \bibliographystyle{ieeenat_fullname}
    \bibliography{main}

\begin{thebibliography}{90}
\providecommand{\natexlab}[1]{#1}
\providecommand{\url}[1]{\texttt{#1}}
\expandafter\ifx\csname urlstyle\endcsname\relax
  \providecommand{\doi}[1]{doi: #1}\else
  \providecommand{\doi}{doi: \begingroup \urlstyle{rm}\Url}\fi

\bibitem[Wang et~al.(2022)Wang, Rong, Liu, Yan, Lin, and Dai]{wang2022towards}
Jingbo Wang, Yu Rong, Jingyuan Liu, Sijie Yan, Dahua Lin, and Bo Dai.
\newblock Towards diverse and natural scene-aware 3d human motion synthesis.
\newblock In \emph{CVPR}, pages 20428--20437, 2022.

\bibitem[Xiao et~al.(2024)Xiao, Wang, Wang, Cao, Zhang, Dai, Lin, and Pang]{xiao2024unified}
Zeqi Xiao, Tai Wang, Jingbo Wang, Jinkun Cao, Wenwei Zhang, Bo Dai, Dahua Lin, and Jiangmiao Pang.
\newblock Unified human-scene interaction via prompted chain-of-contacts.
\newblock In \emph{ICLR}, 2024.

\bibitem[Guo et~al.(2022)Guo, Zuo, Wang, and Cheng]{tm2t}
Chuan Guo, Xinxin Zuo, Sen Wang, and Li Cheng.
\newblock Tm2t: Stochastic and tokenized modeling for the reciprocal generation of 3d human motions and texts.
\newblock In \emph{ECCV}, pages 580--597, 2022.

\bibitem[Jiang et~al.(2024)Jiang, Chen, Liu, Yu, Yu, and Chen]{motiongpt}
Biao Jiang, Xin Chen, Wen Liu, Jingyi Yu, Gang Yu, and Tao Chen.
\newblock Motiongpt: Human motion as a foreign language.
\newblock \emph{NeurIPS}, 2024.

\bibitem[Pan et~al.(2024)Pan, Wang, Huang, Zhang, Wang, Tang, and Wang]{pan2023synthesizing}
Liang Pan, Jingbo Wang, Buzhen Huang, Junyu Zhang, Haofan Wang, Xu Tang, and Yangang Wang.
\newblock Synthesizing physically plausible human motions in 3d scenes.
\newblock In \emph{3DV}, 2024.

\bibitem[Wang et~al.(2023)Wang, Yuan, Luo, Xie, Lin, Iqbal, Fidler, and Khamis]{wang2023learning}
Jingbo Wang, Ye Yuan, Zhengyi Luo, Kevin Xie, Dahua Lin, Umar Iqbal, Sanja Fidler, and Sameh Khamis.
\newblock Learning human dynamics in autonomous driving scenarios.
\newblock In \emph{ICCV}, pages 20739--20749, 2023.

\bibitem[Tevet et~al.(2022)Tevet, Gordon, Hertz, Bermano, and Cohen-Or]{motionclip}
Guy Tevet, Brian Gordon, Amir Hertz, Amit~H Bermano, and Daniel Cohen-Or.
\newblock Motionclip: Exposing human motion generation to clip space.
\newblock In \emph{ECCV}, pages 358--374, 2022.

\bibitem[Petrovich et~al.(2022)Petrovich, Black, and Varol]{temos}
Mathis Petrovich, Michael~J Black, and G{\"u}l Varol.
\newblock Temos: Generating diverse human motions from textual descriptions.
\newblock In \emph{ECCV}, pages 480--497, 2022.

\bibitem[Zhang et~al.(2024)Zhang, Cai, Pan, Hong, Guo, Yang, and Liu]{motiondiffuse}
Mingyuan Zhang, Zhongang Cai, Liang Pan, Fangzhou Hong, Xinying Guo, Lei Yang, and Ziwei Liu.
\newblock Motiondiffuse: Text-driven human motion generation with diffusion model.
\newblock \emph{IEEE TPAMI}, 2024.

\bibitem[Tevet et~al.(2022)Tevet, Raab, Gordon, Shafir, Cohen-Or, and Bermano]{mdm}
Guy Tevet, Sigal Raab, Brian Gordon, Yonatan Shafir, Daniel Cohen-Or, and Amit~H Bermano.
\newblock Human motion diffusion model.
\newblock In \emph{ICLR}, 2022.

\bibitem[Wang et~al.(2022)Wang, Chen, Liu, Zhu, Liang, and Huang]{humanise}
Zan Wang, Yixin Chen, Tengyu Liu, Yixin Zhu, Wei Liang, and Siyuan Huang.
\newblock Humanise: Language-conditioned human motion generation in 3d scenes.
\newblock \emph{NeurIPS}, pages 14959--14971, 2022.

\bibitem[Chen et~al.(2023)Chen, Jiang, Liu, Huang, Fu, Chen, and Yu]{mld}
Xin Chen, Biao Jiang, Wen Liu, Zilong Huang, Bin Fu, Tao Chen, and Gang Yu.
\newblock Executing your commands via motion diffusion in latent space.
\newblock In \emph{CVPR}, pages 18000--18010, 2023.

\bibitem[Yuan et~al.(2023)Yuan, Song, Iqbal, Vahdat, and Kautz]{physdiff}
Ye Yuan, Jiaming Song, Umar Iqbal, Arash Vahdat, and Jan Kautz.
\newblock Physdiff: Physics-guided human motion diffusion model.
\newblock In \emph{ICCV}, pages 16010--16021, 2023.

\bibitem[Zhang et~al.(2023)Zhang, Zhang, Cun, Zhang, Zhao, Lu, Shen, and Shan]{t2mgpt}
Jianrong Zhang, Yangsong Zhang, Xiaodong Cun, Yong Zhang, Hongwei Zhao, Hongtao Lu, Xi Shen, and Ying Shan.
\newblock Generating human motion from textual descriptions with discrete representations.
\newblock In \emph{CVPR}, pages 14730--14740, 2023.

\bibitem[Shafir et~al.(2024)Shafir, Tevet, Kapon, and Bermano]{diffprior}
Yonatan Shafir, Guy Tevet, Roy Kapon, and Amit~H Bermano.
\newblock Human motion diffusion as a generative prior.
\newblock In \emph{ICLR}, 2024.

\bibitem[Karunratanakul et~al.(2023)Karunratanakul, Preechakul, Suwajanakorn, and Tang]{gmd}
Korrawe Karunratanakul, Konpat Preechakul, Supasorn Suwajanakorn, and Siyu Tang.
\newblock Guided motion diffusion for controllable human motion synthesis.
\newblock In \emph{CVPR}, pages 2151--2162, 2023.

\bibitem[Chen et~al.(2024)Chen, Lu, Dai, Dou, Ju, Wang, Komura, and Zhang]{motionclr}
Ling-Hao Chen, Shunlin Lu, Wenxun Dai, Zhiyang Dou, Xuan Ju, Jingbo Wang, Taku Komura, and Lei Zhang.
\newblock Pay attention and move better: Harnessing attention for interactive motion generation and training-free editing, 2024.

\bibitem[Xiao et~al.(2024)Xiao, Wang, Wang, Cao, Zhang, Dai, Lin, and Pang]{unihsi}
Zeqi Xiao, Tai Wang, Jingbo Wang, Jinkun Cao, Wenwei Zhang, Bo Dai, Dahua Lin, and Jiangmiao Pang.
\newblock Unified human-scene interaction via prompted chain-of-contacts.
\newblock In \emph{ICLR}, 2024.

\bibitem[Xie et~al.(2024)Xie, Jampani, Zhong, Sun, and Jiang]{omnicontrol}
Yiming Xie, Varun Jampani, Lei Zhong, Deqing Sun, and Huaizu Jiang.
\newblock Omnicontrol: Control any joint at any time for human motion generation.
\newblock In \emph{ICLR}, 2024.

\bibitem[Lu et~al.(2024)Lu, Chen, Zeng, Lin, Zhang, Zhang, and Shum]{humantomato}
Shunlin Lu, Ling-Hao Chen, Ailing Zeng, Jing Lin, Ruimao Zhang, Lei Zhang, and Heung-Yeung Shum.
\newblock Humantomato: Text-aligned whole-body motion generation.
\newblock \emph{ICML}, 2024.

\bibitem[Dai et~al.(2024)Dai, Chen, Wang, Liu, Dai, and Tang]{motionlcm}
Wenxun Dai, Ling-Hao Chen, Jingbo Wang, Jinpeng Liu, Bo Dai, and Yansong Tang.
\newblock Motionlcm: Real-time controllable motion generation via latent consistency model.
\newblock \emph{ECCV}, 2024.

\bibitem[Chen et~al.(2023)Chen, Zhang, Li, Pang, Xia, and Liu]{humanmac}
Ling-Hao Chen, Jiawei Zhang, Yewen Li, Yiren Pang, Xiaobo Xia, and Tongliang Liu.
\newblock Humanmac: Masked motion completion for human motion prediction.
\newblock In \emph{ICCV}, pages 9544--9555, 2023.

\bibitem[Lu et~al.(2024)Lu, Wang, Lu, Chen, Dai, Dong, Dou, Dai, and Zhang]{lu2024scamo}
Shunlin Lu, Jingbo Wang, Zeyu Lu, Ling-Hao Chen, Wenxun Dai, Junting Dong, Zhiyang Dou, Bo Dai, and Ruimao Zhang.
\newblock Scamo: Exploring the scaling law in autoregressive motion generation model.
\newblock \emph{arXiv preprint arXiv:2412.14559}, 2024.

\bibitem[Dai et~al.(2025)Dai, Chen, Huo, Wang, Liu, Dai, and Tang]{motionlcmv2}
Wenxun Dai, Ling-Hao Chen, Yufei Huo, Jingbo Wang, Jinpeng Liu, Bo Dai, and Yansong Tang.
\newblock Real-time controllable motion generation via latent consistency model.
\newblock 2025.

\bibitem[Goel et~al.(2023)Goel, Pavlakos, Rajasegaran, Kanazawa*, and Malik*]{goel2023humans}
Shubham Goel, Georgios Pavlakos, Jathushan Rajasegaran, Angjoo Kanazawa*, and Jitendra Malik*.
\newblock Humans in 4{D}: Reconstructing and tracking humans with transformers.
\newblock In \emph{ICCV}, 2023.

\bibitem[Ye et~al.(2023)Ye, Pavlakos, Malik, and Kanazawa]{ye2023decoupling}
Vickie Ye, Georgios Pavlakos, Jitendra Malik, and Angjoo Kanazawa.
\newblock Decoupling human and camera motion from videos in the wild.
\newblock In \emph{CVPR}, 2023.

\bibitem[Wang et~al.(2024)Wang, Wang, Liu, and Daniilidis]{wang2024tram}
Yufu Wang, Ziyun Wang, Lingjie Liu, and Kostas Daniilidis.
\newblock Tram: Global trajectory and motion of 3d humans from in-the-wild videos.
\newblock \emph{ECCV}, 2024.

\bibitem[Shin et~al.(2024)Shin, Kim, Halilaj, and Black]{shin2024wham}
Soyong Shin, Juyong Kim, Eni Halilaj, and Michael~J. Black.
\newblock {WHAM}: Reconstructing world-grounded humans with accurate {3D} motion.
\newblock In \emph{CVPR}, 2024.

\bibitem[Shen et~al.(2024)Shen, Pi, Xia, Cen, Peng, Hu, Bao, Hu, and Zhou]{shen2024gvhmr}
Zehong Shen, Huaijin Pi, Yan Xia, Zhi Cen, Sida Peng, Zechen Hu, Hujun Bao, Ruizhen Hu, and Xiaowei Zhou.
\newblock World-grounded human motion recovery via gravity-view coordinates.
\newblock In \emph{ACM SIGGRAPH Asia}, 2024.

\bibitem[Kanazawa et~al.(2019)Kanazawa, Zhang, Felsen, and Malik]{angjoo2019learning}
Angjoo Kanazawa, Jason~Y. Zhang, Panna Felsen, and Jitendra Malik.
\newblock Learning 3d human dynamics from video.
\newblock In \emph{CVPR}, 2019.

\bibitem[Kocabas et~al.(2024)Kocabas, Yuan, Molchanov, Guo, Black, Hilliges, Kautz, and Iqbal]{kocabas2024pace}
Muhammed Kocabas, Ye Yuan, Pavlo Molchanov, Yunrong Guo, Michael~J. Black, Otmar Hilliges, Jan Kautz, and Umar Iqbal.
\newblock Pace: Human and motion estimation from in-the-wild videos.
\newblock In \emph{3DV}, 2024.

\bibitem[Chen et~al.(2024)Chen, Lu, Zeng, Zhang, Wang, Zhang, and Zhang]{chen2024motionllm}
Ling-Hao Chen, Shunlin Lu, Ailing Zeng, Hao Zhang, Benyou Wang, Ruimao Zhang, and Lei Zhang.
\newblock Motionllm: Understanding human behaviors from human motions and videos.
\newblock \emph{arXiv preprint arXiv:2405.20340}, 2024.

\bibitem[Plappert et~al.(2018)Plappert, Mandery, and Asfour]{plappert2018learning}
Matthias Plappert, Christian Mandery, and Tamim Asfour.
\newblock Learning a bidirectional mapping between human whole-body motion and natural language using deep recurrent neural networks.
\newblock \emph{RAS}, 109:\penalty0 13--26, 2018.

\bibitem[Hong et~al.(2022)Hong, Zhang, Pan, Cai, Yang, and Liu]{avatarclip}
Fangzhou Hong, Mingyuan Zhang, Liang Pan, Zhongang Cai, Lei Yang, and Ziwei Liu.
\newblock Avatarclip: Zero-shot text-driven generation and animation of 3d avatars.
\newblock \emph{ACM SIGGRAPH}, 2022.

\bibitem[Athanasiou et~al.(2022)Athanasiou, Petrovich, Black, and Varol]{teach}
Nikos Athanasiou, Mathis Petrovich, Michael~J Black, and G{\"u}l Varol.
\newblock Teach: Temporal action composition for 3d humans.
\newblock In \emph{3DV}, pages 414--423, 2022.

\bibitem[Dabral et~al.(2023)Dabral, Mughal, Golyanik, and Theobalt]{mofusion}
Rishabh Dabral, Muhammad~Hamza Mughal, Vladislav Golyanik, and Christian Theobalt.
\newblock Mofusion: A framework for denoising-diffusion-based motion synthesis.
\newblock In \emph{CVPR}, pages 9760--9770, 2023.

\bibitem[Zhang et~al.(2023)Zhang, Guo, Pan, Cai, Hong, Li, Yang, and Liu]{remodiffuse}
Mingyuan Zhang, Xinying Guo, Liang Pan, Zhongang Cai, Fangzhou Hong, Huirong Li, Lei Yang, and Ziwei Liu.
\newblock Remodiffuse: Retrieval-augmented motion diffusion model.
\newblock In \emph{ICCV}, 2023.

\bibitem[Wan et~al.(2024)Wan, Dou, Komura, Wang, Jayaraman, and Liu]{tlcontrol}
Weilin Wan, Zhiyang Dou, Taku Komura, Wenping Wang, Dinesh Jayaraman, and Lingjie Liu.
\newblock Tlcontrol: Trajectory and language control for human motion synthesis.
\newblock \emph{ECCV}, 2024.

\bibitem[Guo et~al.(2024)Guo, Mu, Javed, Wang, and Cheng]{momask}
Chuan Guo, Yuxuan Mu, Muhammad~Gohar Javed, Sen Wang, and Li Cheng.
\newblock Momask: Generative masked modeling of 3d human motions.
\newblock In \emph{CVPR}, pages 1900--1910, 2024.

\bibitem[Liu et~al.(2024)Liu, Dai, Wang, Cheng, Tang, and Tong]{promotion}
Jinpeng Liu, Wenxun Dai, Chunyu Wang, Yiji Cheng, Yansong Tang, and Xin Tong.
\newblock Plan, posture and go: Towards open-world text-to-motion generation.
\newblock \emph{ECCV}, 2024.

\bibitem[Han et~al.(2024)Han, Peng, Dong, Ren, Shen, and Xu]{amd}
Bo Han, Hao Peng, Minjing Dong, Yi Ren, Yixuan Shen, and Chang Xu.
\newblock Amd: Autoregressive motion diffusion.
\newblock In \emph{AAAI}, pages 2022--2030, 2024.

\bibitem[Xie et~al.(2024)Xie, Wu, Gao, Sun, Yang, and Liang]{b2ahdm}
Zhenyu Xie, Yang Wu, Xuehao Gao, Zhongqian Sun, Wei Yang, and Xiaodan Liang.
\newblock Towards detailed text-to-motion synthesis via basic-to-advanced hierarchical diffusion model.
\newblock In \emph{AAAI}, pages 6252--6260, 2024.

\bibitem[Zhou et~al.(2024)Zhou, Dou, Cao, Liao, Wang, Wang, Liu, Komura, Wang, and Liu]{emdm}
Wenyang Zhou, Zhiyang Dou, Zeyu Cao, Zhouyingcheng Liao, Jingbo Wang, Wenjia Wang, Yuan Liu, Taku Komura, Wenping Wang, and Lingjie Liu.
\newblock Emdm: Efficient motion diffusion model for fast, high-quality motion generation.
\newblock \emph{ECCV}, 2024.

\bibitem[Petrovich et~al.(2024)Petrovich, Litany, Iqbal, Black, Varol, Bin~Peng, and Rempe]{stmc}
Mathis Petrovich, Or Litany, Umar Iqbal, Michael~J Black, Gul Varol, Xue Bin~Peng, and Davis Rempe.
\newblock Multi-track timeline control for text-driven 3d human motion generation.
\newblock In \emph{CVPRW}, pages 1911--1921, 2024.

\bibitem[Hu et~al.(2025)Hu, Xia, Zhao, and Wu]{hu2025mona}
Boxun Hu, Mingze Xia, Ding Zhao, and Guanlin Wu.
\newblock Mona: Moving object detection from videos shot by dynamic camera, 2025.

\bibitem[Wang et~al.(2024)Wang, Chen, Jia, Li, Zhang, Zhang, Liu, Zhu, Liang, and Huang]{move}
Zan Wang, Yixin Chen, Baoxiong Jia, Puhao Li, Jinlu Zhang, Jingze Zhang, Tengyu Liu, Yixin Zhu, Wei Liang, and Siyuan Huang.
\newblock Move as you say interact as you can: Language-guided human motion generation with scene affordance.
\newblock In \emph{CVPR}, pages 433--444, 2024.

\bibitem[Qing et~al.(2023)Qing, Cai, Yang, and Yang]{qing2023storytomotion}
Zhongfei Qing, Zhongang Cai, Zhitao Yang, and Lei Yang.
\newblock Story-to-motion: Synthesizing infinite and controllable character animation from long text, 2023.

\bibitem[Pang et~al.(2022)Pang, Cai, Yang, Zhang, and Liu]{pang2022benchmarking}
Hui~En Pang, Zhongang Cai, Lei Yang, Tianwei Zhang, and Ziwei Liu.
\newblock Benchmarking and analyzing 3d human pose and shape estimation beyond algorithms.
\newblock In \emph{NeurIPS}, 2022.

\bibitem[Moon et~al.(2022)Moon, Choi, and Lee]{moon2022neuralannot}
Gyeongsik Moon, Hongsuk Choi, and Kyoung~Mu Lee.
\newblock Neuralannot: Neural annotator for 3d human mesh training sets.
\newblock In \emph{CVPRW}, 2022.

\bibitem[Moon et~al.(2023)Moon, Choi, Chun, Lee, and Yun]{moon2023three}
Gyeongsik Moon, Hongsuk Choi, Sanghyuk Chun, Jiyoung Lee, and Sangdoo Yun.
\newblock Three recipes for better 3d pseudo-gts of 3d human mesh estimation in the wild.
\newblock In \emph{CVPRW}, 2023.

\bibitem[Yi et~al.(2023)Yi, Liang, Liu, Cao, Wen, Bolkart, Tao, and Black]{yi2023generating}
Hongwei Yi, Hualin Liang, Yifei Liu, Qiong Cao, Yandong Wen, Timo Bolkart, Dacheng Tao, and Michael~J Black.
\newblock Generating holistic 3d human motion from speech.
\newblock In \emph{CVPR}, pages 469--480, 2023.

\bibitem[Bowen and Thompson(2013)]{bowen2013grammar}
C.J. Bowen and R. Thompson.
\newblock \emph{Grammar of the Edit}.
\newblock Focal Press, 2013.

\bibitem[Petrovich et~al.(2024)Petrovich, Litany, Iqbal, Black, Varol, Peng, and Rempe]{petrovich2024multitrack}
Mathis Petrovich, Or Litany, Umar Iqbal, Michael~J. Black, G{\"u}l Varol, Xue~Bin Peng, and Davis Rempe.
\newblock Multi-track timeline control for text-driven 3d human motion generation.
\newblock In \emph{CVPRW}, 2024.

\bibitem[Lin et~al.(2023)Lin, Zeng, Lu, Cai, Zhang, Wang, and Zhang]{lin2023motionx}
Jing Lin, Ailing Zeng, Shunlin Lu, Yuanhao Cai, Ruimao Zhang, Haoqian Wang, and Lei Zhang.
\newblock Motion-x: A large-scale 3d expressive whole-body human motion dataset.
\newblock \emph{NeurIPS}, 2023.

\bibitem[Guo et~al.(2022)Guo, Zou, Zuo, Wang, Ji, Li, and Cheng]{humanml3d}
Chuan Guo, Shihao Zou, Xinxin Zuo, Sen Wang, Wei Ji, Xingyu Li, and Li Cheng.
\newblock Generating diverse and natural 3d human motions from text.
\newblock In \emph{CVPR}, pages 5152--5161, 2022.

\bibitem[Pavlakos et~al.(2022)Pavlakos, Malik, and Kanazawa]{pavlakos2022multishot}
Georgios Pavlakos, Jitendra Malik, and Angjoo Kanazawa.
\newblock Human mesh recovery from multiple shots.
\newblock In \emph{CVPR}, 2022.

\bibitem[Wu et~al.(2023)Wu, Lu, Shen, and Yin]{wu2023clipfusion}
Peng Wu, Xiankai Lu, Jianbing Shen, and Yilong Yin.
\newblock Clip fusion with bi-level optimization for human mesh reconstruction from monocular videos.
\newblock In \emph{ACM MM}, page 105–115, New York, NY, USA, 2023. Association for Computing Machinery.

\bibitem[Wang et~al.(2023)Wang, Weng, Xenochristou, Araujo, Gu, Liu, and Yeung]{wang2022nemo}
Kuan-Chieh Wang, Zhenzhen Weng, Maria Xenochristou, Joao~Pedro Araujo, Jeffrey Gu, C~Karen Liu, and Serena Yeung.
\newblock Nemo: 3d neural motion fields from multiple video instances of the same action.
\newblock In \emph{CVPR}, 2023.

\bibitem[Baradel et~al.(2021)Baradel, Groueix, Weinzaepfel, Brégier, Kalantidis, and Rogez]{baradel2021leveraging}
Fabien Baradel, Thibault Groueix, Philippe Weinzaepfel, Romain Brégier, Yannis Kalantidis, and Grégory Rogez.
\newblock Leveraging mocap data for human mesh recovery.
\newblock In \emph{3DV}, pages 586--595, 2021.

\bibitem[Li et~al.(2021)Li, Yang, Ross, and Kanazawa]{li2021aistpp}
Ruilong Li, Shan Yang, David~A. Ross, and Angjoo Kanazawa.
\newblock Ai choreographer: Music conditioned 3d dance generation with aist++, 2021.

\bibitem[Ionescu et~al.(2014)Ionescu, Papava, Olaru, and Sminchisescu]{ionescu2014human36m}
Catalin Ionescu, Dragos Papava, Vlad Olaru, and Cristian Sminchisescu.
\newblock Human3.6m: Large scale datasets and predictive methods for 3d human sensing in natural environments.
\newblock \emph{IEEE TPAMI}, 36\penalty0 (7):\penalty0 1325--1339, 2014.

\bibitem[Romero et~al.(2017)Romero, Tzionas, and Black]{romero2017mano}
Javier Romero, Dimitrios Tzionas, and Michael~J. Black.
\newblock Embodied hands: Modeling and capturing hands and bodies together.
\newblock \emph{ACM TOG}, 36\penalty0 (6), 2017.

\bibitem[Bogo et~al.(2016)Bogo, Kanazawa, Lassner, Gehler, Romero, and Black]{bogo2016smplify}
Federica Bogo, Angjoo Kanazawa, Christoph Lassner, Peter Gehler, Javier Romero, and Michael~J. Black.
\newblock Keep it {SMPL}: Automatic estimation of {3D} human pose and shape from a single image.
\newblock In \emph{Computer Vision -- ECCV 2016}. Springer International Publishing, 2016.

\bibitem[Arnab et~al.(2019)Arnab, Doersch, and Zisserman]{arnab2019exploiting}
Anurag* Arnab, Carl* Doersch, and Andrew Zisserman.
\newblock Exploiting temporal context for 3d human pose estimation in the wild.
\newblock In \emph{CVPR}, 2019.

\bibitem[Osman et~al.(2020)Osman, Bolkart, and Black]{osman2020star}
Ahmed A.~A. Osman, Timo Bolkart, and Michael~J. Black.
\newblock \emph{STAR: Sparse Trained Articulated Human Body Regressor}, page 598–613.
\newblock Springer International Publishing, 2020.

\bibitem[Huang et~al.(2017)Huang, Bogo, Lassner, Kanazawa, Gehler, Romero, Akhter, and Black]{huang2017towards}
Yinghao Huang, Federica Bogo, Christoph Lassner, Angjoo Kanazawa, Peter~V. Gehler, Javier Romero, Ijaz Akhter, and Michael~J. Black.
\newblock Towards accurate marker-less human shape and pose estimation over time.
\newblock In \emph{3DV}, 2017.

\bibitem[Kanazawa et~al.(2018)Kanazawa, Black, Jacobs, and Malik]{kanazawa2017end}
Angjoo Kanazawa, Michael~J. Black, David~W. Jacobs, and Jitendra Malik.
\newblock End-to-end recovery of human shape and pose.
\newblock In \emph{CVPR}, 2018.

\bibitem[Kocabas et~al.(2020)Kocabas, Athanasiou, and Black]{kocabas2019vibe}
Muhammed Kocabas, Nikos Athanasiou, and Michael~J. Black.
\newblock Vibe: Video inference for human body pose and shape estimation.
\newblock In \emph{CVPR}, 2020.

\bibitem[Kolotouros et~al.(2019)Kolotouros, Pavlakos, and Daniilidis]{kolotouros2019cmr}
Nikos Kolotouros, Georgios Pavlakos, and Kostas Daniilidis.
\newblock Convolutional mesh regression for single-image human shape reconstruction.
\newblock In \emph{CVPR}, 2019.

\bibitem[Luo et~al.(2020)Luo, Golestaneh, and Kitani]{luo20203dhuman}
Zhengyi Luo, S.~Alireza Golestaneh, and Kris~M. Kitani.
\newblock 3d human motion estimation via motion compression and refinement.
\newblock In \emph{ACCV}, 2020.

\bibitem[Vaswani et~al.(2017)Vaswani, Shazeer, Parmar, Uszkoreit, Jones, Gomez, Kaiser, and Polosukhin]{vaswani2023attention}
Ashish Vaswani, Noam Shazeer, Niki Parmar, Jakob Uszkoreit, Llion Jones, Aidan~N Gomez, \L~ukasz Kaiser, and Illia Polosukhin.
\newblock Attention is all you need.
\newblock In \emph{NeurIPS}. Curran Associates, Inc., 2017.

\bibitem[Huang et~al.(2022)Huang, Yi, H{\"o}schle, Safroshkin, Alexiadis, Polikovsky, Scharstein, and Black]{huang2022rich}
Chun-Hao~P. Huang, Hongwei Yi, Markus H{\"o}schle, Matvey Safroshkin, Tsvetelina Alexiadis, Senya Polikovsky, Daniel Scharstein, and Michael~J. Black.
\newblock Capturing and inferring dense full-body human-scene contact.
\newblock In \emph{CVPR}, pages 13274--13285, 2022.

\bibitem[Kaufmann et~al.(2023)Kaufmann, Song, Guo, Shen, Jiang, Tang, Z{\'a}rate, and Hilliges]{kaufmann2023emdb}
Manuel Kaufmann, Jie Song, Chen Guo, Kaiyue Shen, Tianjian Jiang, Chengcheng Tang, Juan~Jos{\'e} Z{\'a}rate, and Otmar Hilliges.
\newblock {EMDB}: The {E}lectromagnetic {D}atabase of {G}lobal 3{D} {H}uman {P}ose and {S}hape in the {W}ild.
\newblock In \emph{ICCV}, 2023.

\bibitem[Yi et~al.(2021)Yi, Zhou, and Xu]{yi2021transpose}
Xinyu Yi, Yuxiao Zhou, and Feng Xu.
\newblock Transpose: Real-time 3d human translation and pose estimation with six inertial sensors.
\newblock \emph{ACM TOG}, 40\penalty0 (4), 2021.

\bibitem[Teed and Deng(2021)]{teed2021droid}
Zachary Teed and Jia Deng.
\newblock Droid-slam: Deep visual slam for monocular, stereo, and rgb-d cameras.
\newblock In \emph{NeurIPS}, pages 16558--16569. Curran Associates, Inc., 2021.

\bibitem[Teed et~al.(2023)Teed, Lipson, and Deng]{teed2023deep}
Zachary Teed, Lahav Lipson, and Jia Deng.
\newblock Deep patch visual odometry.
\newblock \emph{NeurIPS}, 2023.

\bibitem[Chen et~al.(2024)Chen, Chen, Wang, and Pollefeys]{chen2024leap}
Weirong Chen, Le Chen, Rui Wang, and Marc Pollefeys.
\newblock Leap-vo: Long-term effective any point tracking for visual odometry.
\newblock In \emph{CVPR}, 2024.

\bibitem[Mahmood et~al.(2019)Mahmood, Ghorbani, Troje, Pons-Moll, and Black]{mahmood2019amass}
Naureen Mahmood, Nima Ghorbani, Nikolaus~F. Troje, Gerard Pons-Moll, and Michael~J. Black.
\newblock {AMASS}: Archive of motion capture as surface shapes.
\newblock In \emph{ICCV}, pages 5442--5451, 2019.

\bibitem[von Marcard et~al.(2018)von Marcard, Henschel, Black, Rosenhahn, and Pons-Moll]{von20183dpw}
Timo von Marcard, Roberto Henschel, Michael Black, Bodo Rosenhahn, and Gerard Pons-Moll.
\newblock Recovering accurate 3d human pose in the wild using imus and a moving camera.
\newblock In \emph{ECCV}, 2018.

\bibitem[Xu et~al.(2022)Xu, Zhang, Zhang, and Tao]{xu2022vitpose}
Yufei Xu, Jing Zhang, Qiming Zhang, and Dacheng Tao.
\newblock Vi{TP}ose: Simple vision transformer baselines for human pose estimation.
\newblock In \emph{NeurIPS}, 2022.

\bibitem[Dosovitskiy et~al.(2021)Dosovitskiy, Beyer, Kolesnikov, Weissenborn, Zhai, Unterthiner, Dehghani, Minderer, Heigold, Gelly, Uszkoreit, and Houlsby]{dosovitskiy2020vit}
Alexey Dosovitskiy, Lucas Beyer, Alexander Kolesnikov, Dirk Weissenborn, Xiaohua Zhai, Thomas Unterthiner, Mostafa Dehghani, Matthias Minderer, Georg Heigold, Sylvain Gelly, Jakob Uszkoreit, and Neil Houlsby.
\newblock An image is worth 16x16 words: Transformers for image recognition at scale.
\newblock \emph{ICLR}, 2021.

\bibitem[Loper et~al.(2015)Loper, Mahmood, Romero, Pons-Moll, and Black]{loper2015smpl}
Matthew Loper, Naureen Mahmood, Javier Romero, Gerard Pons-Moll, and Michael~J. Black.
\newblock {SMPL}: A skinned multi-person linear model.
\newblock \emph{ACM TOG}, 34\penalty0 (6):\penalty0 248:1--248:16, 2015.

\bibitem[Huang et~al.(2020)Huang, Xiong, Rao, Wang, and Lin]{huang2020movienet}
Qingqiu Huang, Yu Xiong, Anyi Rao, Jiaze Wang, and Dahua Lin.
\newblock Movienet: A holistic dataset for movie understanding.
\newblock In \emph{ECCV}, 2020.

\bibitem[Contributors(2020)]{mmtrack2020}
MMTracking Contributors.
\newblock {MMTracking: OpenMMLab} video perception toolbox and benchmark.
\newblock \url{https://github.com/open-mmlab/mmtracking}, 2020.

\bibitem[Karaev et~al.(2024)Karaev, Rocco, Graham, Neverova, Vedaldi, and Rupprecht]{karaev2023cotracker}
Nikita Karaev, Ignacio Rocco, Benjamin Graham, Natalia Neverova, Andrea Vedaldi, and Christian Rupprecht.
\newblock {CoTracker}: It is better to track together.
\newblock In \emph{ECCV}, 2024.

\bibitem[Kirillov et~al.(2023)Kirillov, Mintun, Ravi, Mao, Rolland, Gustafson, Xiao, Whitehead, Berg, Lo, Doll{\'a}r, and Girshick]{kirillov2023segany}
Alexander Kirillov, Eric Mintun, Nikhila Ravi, Hanzi Mao, Chloe Rolland, Laura Gustafson, Tete Xiao, Spencer Whitehead, Alexander~C. Berg, Wan-Yen Lo, Piotr Doll{\'a}r, and Ross Girshick.
\newblock Segment anything.
\newblock In \emph{ICCV}, 2023.

\bibitem[Fischler and Bolles(1981)]{fischler1981ransac}
Martin~A. Fischler and Robert~C. Bolles.
\newblock Random sample consensus: a paradigm for model fitting with applications to image analysis and automated cartography.
\newblock \emph{Commun. ACM}, 24\penalty0 (6):\penalty0 381–395, 1981.

\bibitem[Yang et~al.(2023)Yang, Zeng, Liu, Li, Zhang, and Zhang]{yang2023explicit}
Jie Yang, Ailing Zeng, Shilong Liu, Feng Li, Ruimao Zhang, and Lei Zhang.
\newblock Explicit box detection unifies end-to-end multi-person pose estimation.
\newblock In \emph{ICLR}, 2023.

\bibitem[Hartley and Zisserman(2003)]{hartley2003triangulation}
Richard Hartley and Andrew Zisserman.
\newblock \emph{Multiple View Geometry in Computer Vision}.
\newblock Cambridge University Press, USA, 2 edition, 2003.

\bibitem[Black et~al.(2023)Black, Patel, Tesch, and Yang]{black2023bedlam}
Michael~J. Black, Priyanka Patel, Joachim Tesch, and Jinlong Yang.
\newblock {BEDLAM}: A synthetic dataset of bodies exhibiting detailed lifelike animated motion.
\newblock In \emph{CVPR}, pages 8726--8737, 2023.

\end{thebibliography}
}
\appendix
\clearpage
\etocdepthtag.toc{mtappendix}
\etocsettagdepth{mtchapter}{none}
\etocsettagdepth{mtappendix}{subsection}
\setcounter{page}{1}
\maketitlesupplementary
\onecolumn

\clearpage

\section{Masked DPVO Detailed Algorithm}
In this section, we provide a detailed exposition of the algorithm underpinning our proposed Masked DPVO, as introduced in the context of \textbf{Comparison on Camera Trajectory Estimation} in \cref{sec:ablation_study}, building on the foundation of DPVO~\cite{teed2023deep}. In accordance with the principles of DPVO, a scene is conceptualized as a composition of camera poses $\mathbf{G} \in SE(3)^{N}$ and a collection of square image patches $\mathbf{P}$ derived from video frames. The inverse depth is denoted as $\mathbf{d}$, while pixel coordinates are represented by $(x, y)$. Each patch $\mathbf{P}_k$ is modeled as a $4p^2$ homogeneous array, where $p$ denotes the patch width, as illustrated below,
\begin{align}
    \mathbf{P}_k =
    \begin{pmatrix}
    x \\
    y \\
    1 \\
    \mathbf{d}
    \end{pmatrix},
    \quad
    \mathbf{x}, \mathbf{y}, \mathbf{d} \in \mathbb{R}^{1 \times p^2}.
\end{align}
Let $i, j$ represent the indices corresponding to frame $i$ and frame $j$, respectively. The projection matrix $\mathbf{P_{kj}}$, which maps the patch $\mathbf{P}_k$, extracted from frame $i$, to frame $j$, can then be expressed as follows,
\begin{align}
    \mathbf{P}_{kj} \sim \mathbf{K} \mathbf{G}_j \mathbf{G}_i^{-1} \mathbf{K}^{-1} \mathbf{P}_k.
\end{align}
where, $\mathbf{K}$ is a calibration matrix, defined as,
\begin{align}
    \mathbf{K} = 
    \begin{pmatrix}
    f_x & 0 & c_x & 0 \\
    0 & f_y & c_y & 0 \\
    0 & 0 & 1 & 0 \\
    0 & 0 & 0 & 1
    \end{pmatrix}.
\end{align}

In the original DPVO framework, patches are selected randomly, as this approach has been shown to achieve comparable improvements in performance. However, a critical issue arises when a selected patch corresponds to a region on a moving object. In such cases, the estimated camera pose $\mathbf{G}$ becomes inaccurate, negatively impacting the reconstruction of human motion in world coordinates. Given that our primary focus involves human-centric video data, where moving humans often occupy a significant portion of the image, excluding regions corresponding to moving humans presents a straightforward yet effective strategy for improving camera pose estimation accuracy.

To address this, we incorporate SAM~\cite{kirillov2023segany} into the DPVO pipeline. Using the bounding box of a detected human as a prompt, SAM generates a human mask. During the patch extraction process across frames, if a randomly selected patch falls within the human mask, it is excluded and a new patch is selected instead. The resulting patch after applying the human mask is denoted as $\mathbf{\hat{P}}_k$. The corresponding masked projection matrix is subsequently represented as $\mathbf{P}_{kj}$,
\begin{align}
    \mathbf{\hat{P}}_{kj} \sim \mathbf{K} \mathbf{G}_j \mathbf{G}_i^{-1} \mathbf{K}^{-1} \mathbf{\hat{P}}_k.
\end{align}

Following the principles of DPVO, a bipartite patch graph is constructed to capture the relationships between patches and video frames. In this graph, edges connect patches and frames that are within a specified distance, defined by the temporal relationship between frame $i$ and frame $j$). The graph is inherently dynamic, as older frames are discarded as newer frames are introduced. For each edge ($k$, $j$) in the patch graph, the visual alignment, determined by the current estimates of depth and camera poses, is evaluated. This is achieved by computing correlation features $\mathbf{C} \in \mathtt{R}^{p\times p \times 7 \times 7}$ (visual similarities), which represent visual similarities. These correlation features are computed as follows,
\begin{align}
    \mathbf{C}_{uv\alpha \beta} = \mathbf{g}_{uv} \cdot \mathbf{f}(\mathbf{\hat{P}}_{kj}(u, v) + \Delta_{\alpha\beta})
\end{align}
where $u,v$ are the index of each pixel in patch $k$, $\mathbf{f(\cdot)}$ is the bilinear sampling and $\Delta$ is $7 \times 7$ grid indexed by $\alpha$ and $\beta$. Based on extracted features and correlations, DPVO uses MLP layers to predict trajectory update $\delta_{kj}$ and confidence weight $\Sigma_{kj}$.

Then, a differentiable bundle adjustment (BA) layers is proposed to solve both the depth and camera poses. BA operates on the patch graph and outputs updates to depth and camera pose. The optimization objective of BA is shown as,
\begin{align}
    \sum_{(k,j) \in \mathcal{E}} 
    \left\|
    \mathbf{K} \mathbf{G}_j \mathbf{G}_i^{-1} \mathbf{K}^{-1} \mathbf{\hat{P}}_k - \left[\mathbf{\hat{P}^{'}}_{kj} + \delta_{kj}\right]
    \right\|_{\Sigma_{kj}}^2
\end{align}
where $\left\| \cdot \right\|_{\Sigma}$ is the Mahalanobis distance and $\mathbf{\hat{P}^{'}}_{kj}$ is the center of $\mathbf{\hat{P}}_{kj}$. DPVO then applies two Gauss-Newton iterations to the linearized objective, optimizing the camera poses and inverse depth component at the same time. The main intuition for this optimization is to refine the camera poses and depth such that the induced trajectory updates agree with the predicted trajectory updates.

\clearpage

\section{Masked LEAP-VO Detailed Algorithm}
In this section, we illustrate the detailed algorithm of our proposed camera estimation method Masked LEAP-VO. We start with the preliminaries about tracking any point (TAP), which has been used in LEAP-VO~\cite{chen2024leap}. Given an input video sequence $\mathbf{I} = \{I_t\}_{t=1}^{T}$ of length $T$, where $I_t$ denotes the $t$-th frame, the goal of TAP is to track a set of query points across these frames. For a specific query point $q$ with the 2D pixel coordinate $x_t^{q} \in \mathbb{R}^2$ in frame $I_t$, TAP predicts both the visibility $\mathbf{V}^{q} = [v_1^{q}, \cdots, v_T^{q}]$ and trajectory $\mathbf{X}^{q} = [x_1^{q}, \cdots, x_T^{q}]$ of this point through the whole video. Thus, we have,
\begin{align}
    (\mathbf{X}, \mathbf{V}) = \text{TAP}(\mathbf{I}, x_t^{q}, I_t)
\end{align}
LEAP-VO uses CoTracker~\cite{karaev2023cotracker} to as their TAP module. During inference, CoTracker will produce point feature $f_t$ for each frame $I_t$ and then concatenate these point features into a tensor $\mathbf{F}^{q} = [f_1, \cdots, f_T]$. With predicted $\mathbf{X}$ and $\mathbf{F}$, LEAP-VO tries to predict whether each point query is on the dynamic object (dynamic label $m_d$) using anchor-based trajectory comparison to alleviate the negative effect they bring to the camera pose estimation.

Subsequently, LEAP-VO uses temporal probability modeling to get the confidence of estimated trajectories of each point query. Let $X=[\mathbf{a},\mathbf{b}]$, where $\mathbf{a}, \mathbf{b}$ represent the $x$ and $y$ coordinates of all points in $X$, respectively. The probability density function for a coordinate is modeled using a multivariate Cauchy distribution,
\begin{align}
    p(\mathbf{a}|I_t, &x_i) = \frac{\gamma(\frac{1+T}{2})}{\gamma(\frac{1}{2}) \pi^{\frac{T}{2}}|\sum_a|^{\frac{1}{2}}[1\!+\!(\mathbf{a} \! - \!\mu_a)^T \Sigma_a^{-1} (\mathbf{a} \!- \!\mu_a)]^{\frac{1+S}{2}}},
    \label{eq:leapvo_distribution}
\end{align}
and similarly for $p(\mathbf{b}|I_t, x_i)$. Here, $\gamma$ denotes the Gamma function. LEAPVO then filters out trajectories shorter than a predefined threshold and inputs the remaining trajectories into a bundle adjustment (BA) stage with sliding window optimization. 

However, as we mentioned in \cref{sec:initialization}, the performance of this method is still satisfactory as the exclusion of the moving object is not complete. Since we are dealing with human-centric videos, we can simply apply SAM to get the mask of human and exclude these points. Thus, we set visibility of trajectories that contains points inside human mask to zero. The adjusted BA function is shown as \cref{eq:leapvo_ba} in \cref{sec:initialization}.

\clearpage

\section{Camera Calibration Procedure}
In this section, we show the detailed procedure of the \textit{camera calibration} we mentioned in \cref{sec:orientation_alignment}. We denote the selected feature points of two frames during shot transition in world coordinate as 
\begin{align}
    W_1 = [(x_{w1}^{(1)}, y_{w1}^{(1)}, z_{w1}^{(1)}), (x_{w1}^{(2)}, (y_{w1}^{(2)}, z_{w1}^{(2)}), \cdots, (x_{w1}^{(N)}, y_{w1}^{(N)}, z_{w1}^{(N)})]^{\top}, \notag\\
    W_2 = [(x_{w2}^{(1)}, y_{w2}^{(1)}, z_{w2}^{(1)}), (x_{w2}^{(2)},(y_{w2}^{(2)}, z_{w2}^{(2)}), \cdots, (x_{w2}^{(N)}, y_{w2}^{(N)}, z_{w2}^{(N)})]^{\top}, \notag
\end{align}
respectively. Suppose we have the initialized camera translation matrix $\mathbf{T}$ and camera rotation matrix $\mathbf{R}$ during shot transition from Masked LEAP-VO in \cref{sec:initialization}, where 
\begin{align}
    \mathbf{R} = \begin{bmatrix}
        r_{11} & r_{12} & r_{13} \\
        r_{21} & r_{22} & r_{23} \\
        r_{31} & r_{32} & r_{33}
        \end{bmatrix}
    , \mathbf{T} = \begin{bmatrix} t_x \\ t_y \\ t_z \end{bmatrix},
\end{align}
and since 
\begin{align}
    \begin{bmatrix}
        x_{w1}^{(1)} \\
        y_{w1}^{(1)} \\
        z_{w1}^{(1)}
        \end{bmatrix}
        =
        \begin{bmatrix}
        r_{11} & r_{12} & r_{13} \\
        r_{21} & r_{22} & r_{23} \\
        r_{31} & r_{32} & r_{33}
        \end{bmatrix}
        \begin{bmatrix}
        x_{w1}^{(1)} \\
        y_{w1}^{(1)} \\
        z_{w1}^{(1)}
        \end{bmatrix}
        +
        \begin{bmatrix}
        t_x \\
        t_y \\
        t_z
    \end{bmatrix}.
    \label{eq:constraint_1}
\end{align}
Thus, according to Epipolar Constraint,
\begin{align}
    \begin{bmatrix}
        x_{w2}^{(1)} & y_{w2}^{(1)} & z_{w2}^{(1)}
    \end{bmatrix}
        \begin{bmatrix}
        0 & -t_z & t_y \\
        t_z & 0 & -t_x \\
        -t_y & t_x & 0
        \end{bmatrix}
        \begin{bmatrix}
        x_{w2}^{(1)} \\
        y_{w2}^{(1)} \\
        z_{w2}^{(1)}
    \end{bmatrix}
    = 0.
    \label{eq:constraint_epipolar}
\end{align}
Thus, by substituting \cref{eq:constraint_1} to \cref{eq:constraint_epipolar},
\begin{align}
    \begin{bmatrix}
    x_{w2}^{(1)} & y_{w2}^{(1)} & z_{w2}^{(1)}
    \end{bmatrix}
    \begin{bmatrix}
    0 & -t_z & t_y \\
    t_z & 0 & -t_x \\
    -t_y & t_x & 0
    \end{bmatrix}
    \begin{bmatrix}
    r_{11} & r_{12} & r_{13} \\
    r_{21} & r_{22} & r_{23} \\
    r_{31} & r_{32} & r_{33}
    \end{bmatrix}
    \begin{bmatrix} 
        x_{w1}^{(1)} \\
        y_{w1}^{(1)} \\
        z_{w1}^{(1)}
    \end{bmatrix}
    + \begin{bmatrix}
    0 & -t_z & t_y \\
    t_z & 0 & -t_x \\
    -t_y & t_x & 0
    \end{bmatrix}
    \begin{bmatrix}
    t_x \\
    t_y \\
    t_z
    \end{bmatrix}
    = 0.
    \label{eq:epipolar}
\end{align}
Let's denote $[\mathbf{T}]_{\times}$ as 
\begin{align}
    [\mathbf{T}]_{\times} = \begin{bmatrix}
        0 & -t_z & t_y \\
        t_z & 0 & -t_x \\
        -t_y & t_x & 0
    \end{bmatrix}.
    \label{eq:t_times}
\end{align}
As a result, the essential matrix can be denoted as $\mathbf{E} = [\mathbf{T}]_{\times}\mathbf{R}$,
\begin{align}
    \mathbf{E} = \begin{bmatrix}
    e_{11} & e_{12} & e_{13} \\
    e_{21} & e_{22} & e_{23} \\
    e_{31} & e_{32} & e_{33}
    \end{bmatrix}
    =
    \begin{bmatrix}
    0 & -t_z & t_y \\
    t_z & 0 & -t_x \\
    -t_y & t_x & 0
    \end{bmatrix}
    \begin{bmatrix}
    r_{11} & r_{12} & r_{13} \\
    r_{21} & r_{22} & r_{23} \\
    r_{31} & r_{32} & r_{33}
    \end{bmatrix}.
\end{align}
Since
\begin{align}
    &\begin{bmatrix}
    0 & -t_z & t_y \\
    t_z & 0 & -t_x \\
    -t_y & t_x & 0
    \end{bmatrix}
    \begin{bmatrix}
    t_x \\
    t_y \\
    t_z
    \end{bmatrix}
    = 0,
    \label{eq:t_zero}
\end{align}
by constructing essential matrix $\mathbf{E}$ and elimination of the second term, the \cref{eq:epipolar} can then be rewritten as 
\begin{align}
    \begin{bmatrix}
    x_{w2}^{(1)} & y_{w2}^{(1)} & z_{w2}^{(1)}
    \end{bmatrix}
    \begin{bmatrix}
    e_{11} & e_{12} & e_{13} \\
    e_{21} & e_{22} & e_{23} \\
    e_{31} & e_{32} & e_{33}
    \end{bmatrix}
    \begin{bmatrix} 
        x_{w1}^{(1)} \\
        y_{w1}^{(1)} \\
        z_{w1}^{(1)}
    \end{bmatrix}
    = 0
    \label{eq:new_epipolar}
\end{align}
As our method is targeted for the in-the-wild multi-shot videos, we typically do not know the intrinsics for each shot. Thus, we assume that through a multi-shot video, the camera intrinsics are the same. The camera intrinsic matrix $\mathbf{K}$ is denoted as, 
\begin{align}
    \mathbf{K} = 
    \begin{bmatrix}
    f_x & 0 & o_x \\
    0 & f_y & o_y \\
    0 & 0 & 1
    \end{bmatrix}
\end{align}
Similar as mentioned in \cref{sec:orientation_alignment}, we denote the coordinates of detected 2D KPTs of two frames in the shot transition as $\mathbf{S}_1 = [(x_1^{(1)}, y_1^{(1)}), (x_1^{(2)}, y_1^{(2)}), \cdots, (x_1^{(N)}, y_1^{(N)})]^{\top} \in \mathtt{R}^{2\times N}$ and  $\mathbf{S}_2 = [(x_2^{(1)}, y_2^{(1)}), (x_2^{(2)}, y_2^{(2)}), \cdots, (x_2^{(N)}, y_2^{(N)})]^{\top} \in \mathtt{R}^{2\times N}$. Based on intrinsic matrix $\mathbf{K}$, 
\begin{align}
    \begin{bmatrix}
    x_{w2}^{(1)} & y_{w2}^{(1)} & z_{w2}^{(1)}
    \end{bmatrix}
    =
    \begin{bmatrix}
    x_2^{(1)} & y_2^{(1)} & 1
    \end{bmatrix}
    \begin{bmatrix}
    \frac{1}{f_x} & 0 & 0 \\
    -\frac{o_x}{f_x} & \frac{1}{f_y} & 0 \\
    0 & -\frac{o_y}{f_y} & 1
    \end{bmatrix},
    \begin{bmatrix}
    x_{w1}^{(1)} \\
    y_{w1}^{(1)} \\ 
    z_{w2}^{(1)}
    \end{bmatrix}
    =
    \begin{bmatrix}
    \frac{1}{f_x} & 0 & 0 \\
    -\frac{o_x}{f_x} & \frac{1}{f_y} & 0 \\
    0 & -\frac{o_y}{f_y} & 1
    \end{bmatrix} 
    \begin{bmatrix}
    x_1^{(1)} \\ 
    y_1^{(1)} \\
    1
    \end{bmatrix}
    \label{eq:uv_system}
\end{align}
Thus, we can substitute \cref{eq:uv_system} to \cref{eq:new_epipolar},
\begin{align}
    \setlength{\arraycolsep}{1pt}
    \begin{bmatrix}
    x_2^{(1)} & y_2^{(1)} & 1
    \end{bmatrix}
    \begin{bmatrix}
    \frac{1}{f_x} & 0 & 0 \\
    -\frac{o_x}{f_x} & \frac{1}{f_y} & 0 \\
    0 & -\frac{o_y}{f_y} & 1
    \end{bmatrix}
    \begin{bmatrix}
    e_{11} & e_{12} & e_{13} \\
    e_{21} & e_{22} & e_{23} \\
    e_{31} & e_{32} & e_{33}
    \end{bmatrix} 
    \begin{bmatrix}
    \frac{1}{f_x} & 0 & 0 \\
    -\frac{o_x}{f_x} & \frac{1}{f_y} & 0 \\
    0 & -\frac{o_y}{f_y} & 1
    \end{bmatrix} 
    \begin{bmatrix}
    x_1^{(1)} \\ 
    y_1^{(1)} \\
    1
    \end{bmatrix}
    = 0.
    \label{eq:uv_epipolar}
\end{align}
We can then denote fundamental matrix $\mathbf{F}$ as, 
\begin{align}
\setlength{\arraycolsep}{1pt}
    \mathbf{F} =
    \begin{bmatrix}
    f_{11} & f_{12} & f_{13} \\
    f_{21} & f_{22} & f_{23} \\
    f_{31} & f_{32} & f_{33}
    \end{bmatrix} 
    =\begin{bmatrix}
    \frac{1}{f_x} & 0 & 0 \\
    -\frac{o_x}{f_x} & \frac{1}{f_y} & 0 \\
    0 & -\frac{o_y}{f_y} & 1
    \end{bmatrix}
    \begin{bmatrix}
    e_{11} & e_{12} & e_{13} \\
    e_{21} & e_{22} & e_{23} \\
    e_{31} & e_{32} & e_{33}
    \end{bmatrix} 
    \begin{bmatrix}
    \frac{1}{f_x} & 0 & 0 \\
    -\frac{o_x}{f_x} & \frac{1}{f_y} & 0 \\
    0 & -\frac{o_y}{f_y} & 1
    \end{bmatrix},
    \label{eq:fundamental_matrix}
\end{align}
and we can rewrite \cref{eq:uv_epipolar} as, 
\begin{align}
    \begin{bmatrix}
    x_2^{(1)} & y_2^{(1)} & 1
    \end{bmatrix}
    &\begin{bmatrix}
    f_{11} & f_{12} & f_{13} \\
    f_{21} & f_{22} & f_{23} \\
    f_{31} & f_{32} & f_{33}
    \end{bmatrix} 
    \begin{bmatrix}
    x_1^{(1)} \\ 
    y_1^{(1)} \\
    1
    \end{bmatrix}
    = 0.
    \label{eq:uv_epipolar_fundamental}
\end{align}
One clarification is that the fundamental matrix $\mathbf{F}$ here is denoted as essential matrix $\mathbf{E}$ in \cref{eq:e_matrix} in \cref{sec:orientation_alignment}.
After we derive \cref{eq:uv_epipolar_fundamental}, we can expand it as a linear equation, illustrated as, 
\begin{align}
    \left( f_{11} x_1^{(1)} + f_{12} y_1^{(1)} + f_{13} \right) x_2^{(1)} + f_{31} x_1^{(1)} + f_{32} y_1^{(1)} + f_{33} + \left( f_{21} x_1^{(1)} + f_{22} y_1^{(1)} + f_{23} \right) y_2^{(1)} = 0.
\end{align}
This illustrates the case for one correspondence, since we have correspondences, we can build up a linear equation system to solve $\mathbf{F}$. Thus, we create a matrix $\mathbf{A}$ based on $\mathbf{S_1}$ and $\mathbf{S_2}$, shown as, 
\begin{align}
    A = 
    \begin{bmatrix}
    x_2^{(1)}x_1^{(1)} & x_2^{(1)}y_1^{(1)} & x_2^{(1)} & y_2^{(1)}x_1^{(1)} & y_2^{(1)}y_1^{(1)} & y_2^{(1)} & x_1^{(1)} & y_1^{(1)} & 1 \\
    \vdots & \vdots & \vdots & \vdots & \vdots & \vdots & \vdots & \vdots & \vdots \\
    x_2^{(i)}x_1^{(i)} & x_2^{(i)}y_1^{(i)} & x_2^{(i)} & y_2^{(i)}x_1^{(i)} & y_2^{(i)}y_1^{(i)} & y_2^{(i)} & x_1^{(i)} & y_1^{(i)} & 1 \\
    \vdots & \vdots & \vdots & \vdots & \vdots & \vdots & \vdots & \vdots & \vdots \\
    x_2^{(N)}x_1^{(N)} & x_2^{(N)}y_1^{(N)} & x_2^{(N)} & y_2^{(N)}x_1^{(N)} & y_2^{(N)}y_1^{(N)} & y_2^{(N)} & x_1^{(N)} & y_1^{(N)} & 1
    \end{bmatrix}.
\end{align}
Let $f$ denotes the flattened version $\mathbf{F}$, 
%
\begin{align}
    \mathbf{A}f = \begin{bmatrix}
    x_2^{(1)}x_1^{(1)} & x_2^{(1)}y_1^{(1)} & x_2^{(1)} & y_2^{(1)}x_1^{(1)} & y_2^{(1)}y_1^{(1)} & y_2^{(1)} & x_1^{(1)} & y_1^{(1)} & 1 \\
    \vdots & \vdots & \vdots & \vdots & \vdots & \vdots & \vdots & \vdots & \vdots \\
    x_2^{(i)}x_1^{(i)} & x_2^{(i)}y_1^{(i)} & x_2^{(i)} & y_2^{(i)}x_1^{(i)} & y_2^{(i)}y_1^{(i)} & y_2^{(i)} & x_1^{(i)} & y_1^{(i)} & 1 \\
    \vdots & \vdots & \vdots & \vdots & \vdots & \vdots & \vdots & \vdots & \vdots \\
    x_2^{(N)}x_1^{(N)} & x_2^{(N)}y_1^{(N)} & x_2^{(N)} & y_2^{(N)}x_1^{(N)} & y_2^{(N)}y_1^{(N)} & y_2^{(N)} & x_1^{(N)} & y_1^{(N)} & 1
    \end{bmatrix} 
    \begin{bmatrix}
    f_{11} \\
    f_{12} \\
    f_{13} \\
    f_{21} \\
    f_{22} \\
    f_{23} \\
    f_{31} \\
    f_{32} \\ 
    f_{33}
    \end{bmatrix}
    =0.   
    \label{eq:af}
\end{align}
Then, we can solve $f$ by applying singular value decomposition (SVD) on $\mathbf{A}$. Decompose $\mathbf{A}$ into three matrices $\mathbf{A} = U\Sigma V^{\top}$. Substituting into \cref{eq:af}, 
\begin{align}
    U \Sigma V^{\top} f = 0 \notag \\
    U^{\top} U \Sigma V^{\top} f = U^{\top} 0 \notag \\
    \Sigma V^{\top} f = 0.
\end{align}
Since f is in the null space of $\mathbf{A}$, from decomposition, the null space is spanned by the last columns of $V$ corresponding to zero singular values in $\Sigma$. Thus, we can extract the last column of V (denoted as $v_n$) and assign $f = v_n$.
After we get $f$, we can get the fundamental matrix $\mathbf{F}$. We can then derive essential matrix $\mathbf{E}$ from $\mathbf{F}$ shown as, 
\begin{align}
    \mathbf{E} = \mathbf{K}^{\top}\mathbf{F}\mathbf{K} = \begin{bmatrix}
    f_x & 0 & 0 \\
    0 & f_y & 0 \\
    o_x & o_y & 1
    \end{bmatrix}
    \begin{bmatrix}
    f_{11} & f_{12} & f_{13} \\
    f_{21} & f_{22} & f_{23} \\
    f_{31} & f_{32} & f_{33}
    \end{bmatrix}
    \begin{bmatrix}
    f_x & 0 & o_x \\
    0 & f_y & o_y \\
    0 & 0 & 1
    \end{bmatrix}
\end{align}
Then, we can derive camera rotation $\mathbf{R}$ and camera translation $\mathbf{T}$ by applying SVD on essential matrix $\mathbf{E}$, 
\begin{align}
    \mathbf{E} = U \Sigma V^{\top} = \begin{bmatrix}
        u_{11} & u_{12} & u_{13} \\
        u_{21} & u_{22} & u_{23} \\
        u_{31} & u_{32} & u_{33}
        \end{bmatrix}
        \begin{bmatrix}
        \sigma_1 & 0 & 0 \\
        0 & \sigma_2 & 0 \\
        0 & 0 & \sigma_3
        \end{bmatrix}
        \begin{bmatrix}
        v_{11} & v_{12} & v_{13} \\
        v_{21} & v_{22} & v_{23} \\
        v_{31} & v_{32} & v_{33}
        \end{bmatrix}^\top \\
        \mathbf{R} = \begin{bmatrix}
        u_{11} & u_{12} & u_{13} \\
        u_{21} & u_{22} & u_{23} \\
        u_{31} & u_{32} & u_{33}
        \end{bmatrix}
        \begin{bmatrix}
        0 & -1 & 0 \\
        1 & 0 & 0 \\
        0 & 0 & 1
        \end{bmatrix}
        \begin{bmatrix}
        v_{11} & v_{12} & v_{13} \\
        v_{21} & v_{22} & v_{23} \\
        v_{31} & v_{32} & v_{33}
        \end{bmatrix}^\top,
        \mathbf{T} = 
        \begin{bmatrix}
        u_{13} \\
        u_{23} \\
        u_{33}
        \end{bmatrix}
\end{align}
The output $\mathbf{R}$ here is denoted as the $\mathbf{R}_{\delta_{\text{cam}}}$ in \cref{sec:orientation_alignment}.

\clearpage

\section{Implementation Details of \methodname}

\subsection{Training Details}

The \textit{ms}-HMR, the trajectory, and foot sliding refiner are trained on the AMASS~\cite{mahmood2019amass}, 3DPW~\cite{von20183dpw}, Human3.6M~\cite{ionescu2014human36m}, and BEDLAM~\cite{black2023bedlam} datasets, evaluate on EMDB and our \dataname. During training, we introduce random rotational noise (ranging from 0 to 1 radian) along the y-axis to the root orientation $\Gamma$ and random noise to the body pose $\theta$ at random positions to simulate the inaccuracies of pre-estimated human motions caused by shot transitions in multi-shot videos. This strategy enables the network to robustly recover smooth and consistent human motion from noisy initial parameters.
The introduction of these noise perturbations stems from the observation that relying solely on our orientation alignment module may fall short in fully aligning the root pose across different shots in challenging scenarios, particularly when significant angular discrepancies occur during shot transitions. To overcome this limitation, our trainable module is designed not only to align root poses with camera parameters but also to ensure smooth transitions in local poses across shot boundaries. This strategy significantly enhances the robustness of our method in managing abrupt orientation changes caused by multi-shot video transitions.

\subsection{\dataname Benchmark Details}

Our \dataname Benchmark is building on existing multi-view datasets AIST~\cite{li2021aistpp}, H36M~\cite{ionescu2014human36m}. The AIST dataset provides world translation as ground truth. H36M dataset does not include such labels; therefore, we process human world translation from~\cite{shen2024gvhmr} as the ground truth.
For benchmarking, the test results were obtained after training \methodname for 80 epochs on a single NVIDIA-A100 GPU (1.3 days). This computational setup ensures efficient convergence of the model while maintaining a high level of accuracy in human motion recovery.

\subsection{Baseline Setting}

The baseline mentioned in ~\cref{tab:ablation_results} in \cref{sec:ablation_study} is implemented as follows. For the input multi-shot videos, we process them by using our proposed \textit{shot transition detector} and \textit{human and camera parameters initialization} as shown in ~\cref{fig:pipeline} in \cref{sec:method}. Next, we directly concatenate the human translation, root orientation, and body pose based on the relative offsets between frames. This method could achieve global motion recovery in a few scenarios with simple motions. However, our observations reveal that this approach suffers from noticeable issues, such as foot sliding, motion truncation, and motion collapse. As there were no existing methods for global human motion recovery, this approach can serve as our baseline for conducting ablation studies to evaluate the effectiveness of our proposed training and optimization modules.

\clearpage

\section{Visualization of Comparison between Existing Methods}

\subsection{Visualization of Comparison between Existing Methods}

In this section, we present visual comparisons between our proposed method and existing approaches, including SLAHMR~\cite{ye2023decoupling}, GVHMR~\cite{shen2024gvhmr}, as well as the ground truth. These comparisons aim to highlight the advancements achieved by our method in accurately reconstructing human motion. To ensure a comprehensive evaluation, we provide visualizations from multiple viewpoints: a side view (a), an alternative side view (b), a top-down view (c), and reconstructed motion trajectories (d). The side view (a) allows for a detailed examination of the overall pose accuracy and alignment over time, emphasizing the consistency and anatomical plausibility of the reconstructed movements. The alternative side view (b) provides additional insights into the depth and spatial relationships of the poses, capturing nuances that might be less apparent from a single perspective. The top-down view (c) reveals the positional alignment and the spatial coherence of the trajectories, showcasing the robustness of our approach in reconstructing dynamic and complex motions. Finally, the motion trajectories (d) offer a quantitative and qualitative representation of the movement paths, highlighting the differences between methods and their ability to accurately track and reproduce the ground truth trajectories. These visual comparisons underscore our method ability in delivering precise, consistent, and realistic human motion recovery from multi-shot videos.
\begin{figure*}[!ht]
    \vspace{-0.15in}
    \centering
    \includegraphics[width=0.98\linewidth]{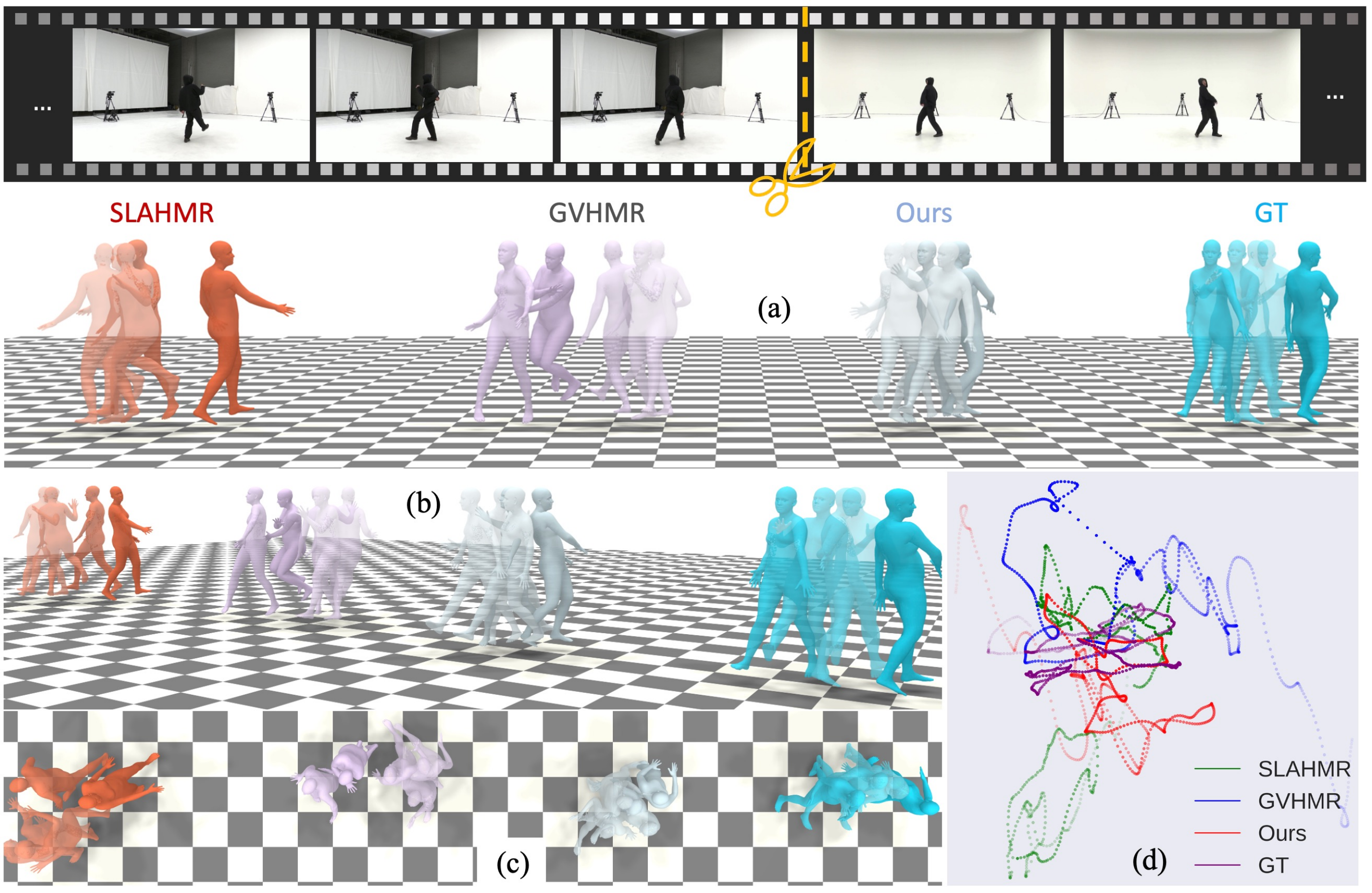}
    \label{fig:vis_comp_1}
\end{figure*}
%
%
\begin{figure*}[!ht]
    \centering
    \includegraphics[width=0.98\linewidth]{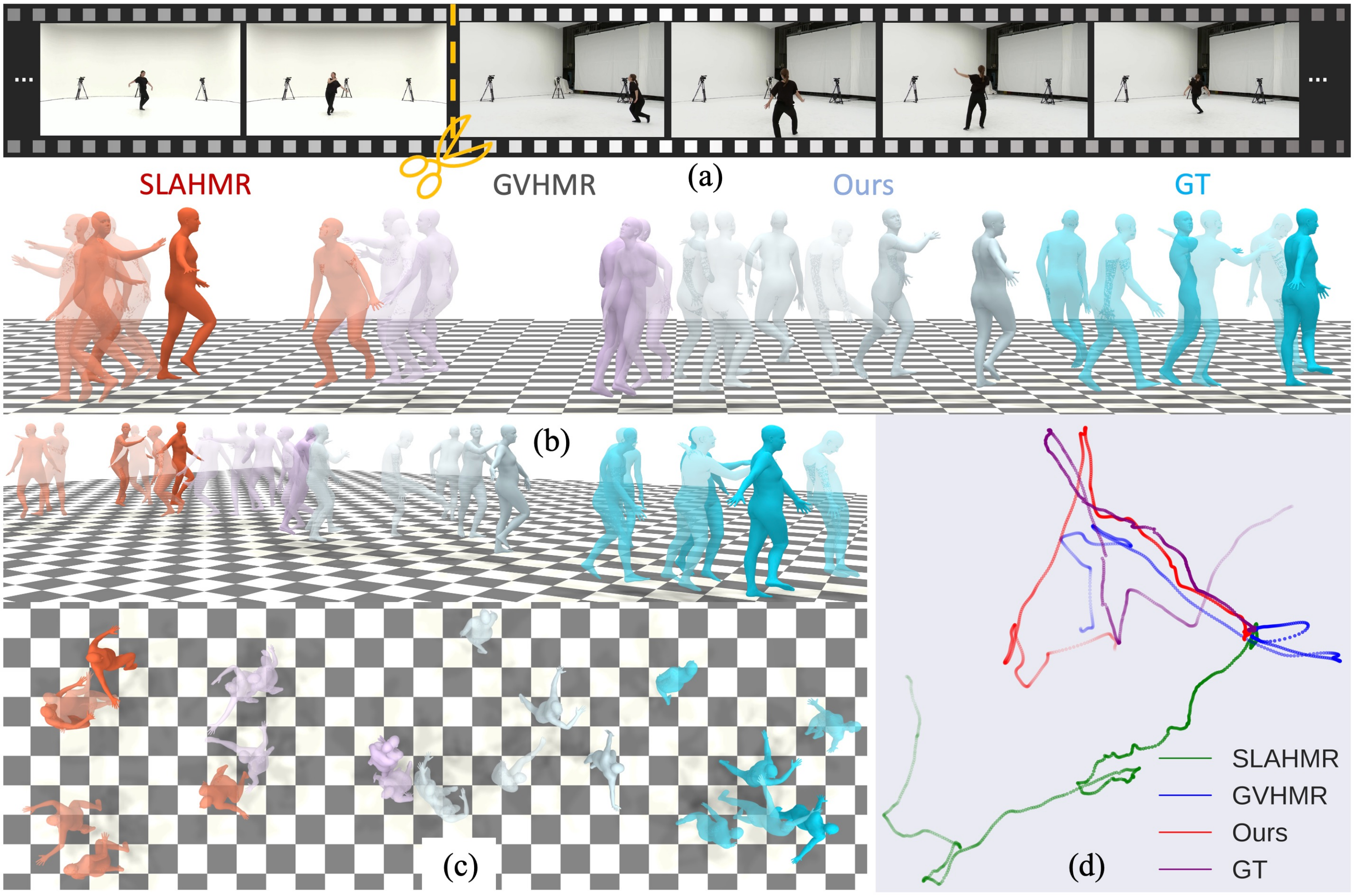}
    \includegraphics[width=0.98\linewidth]{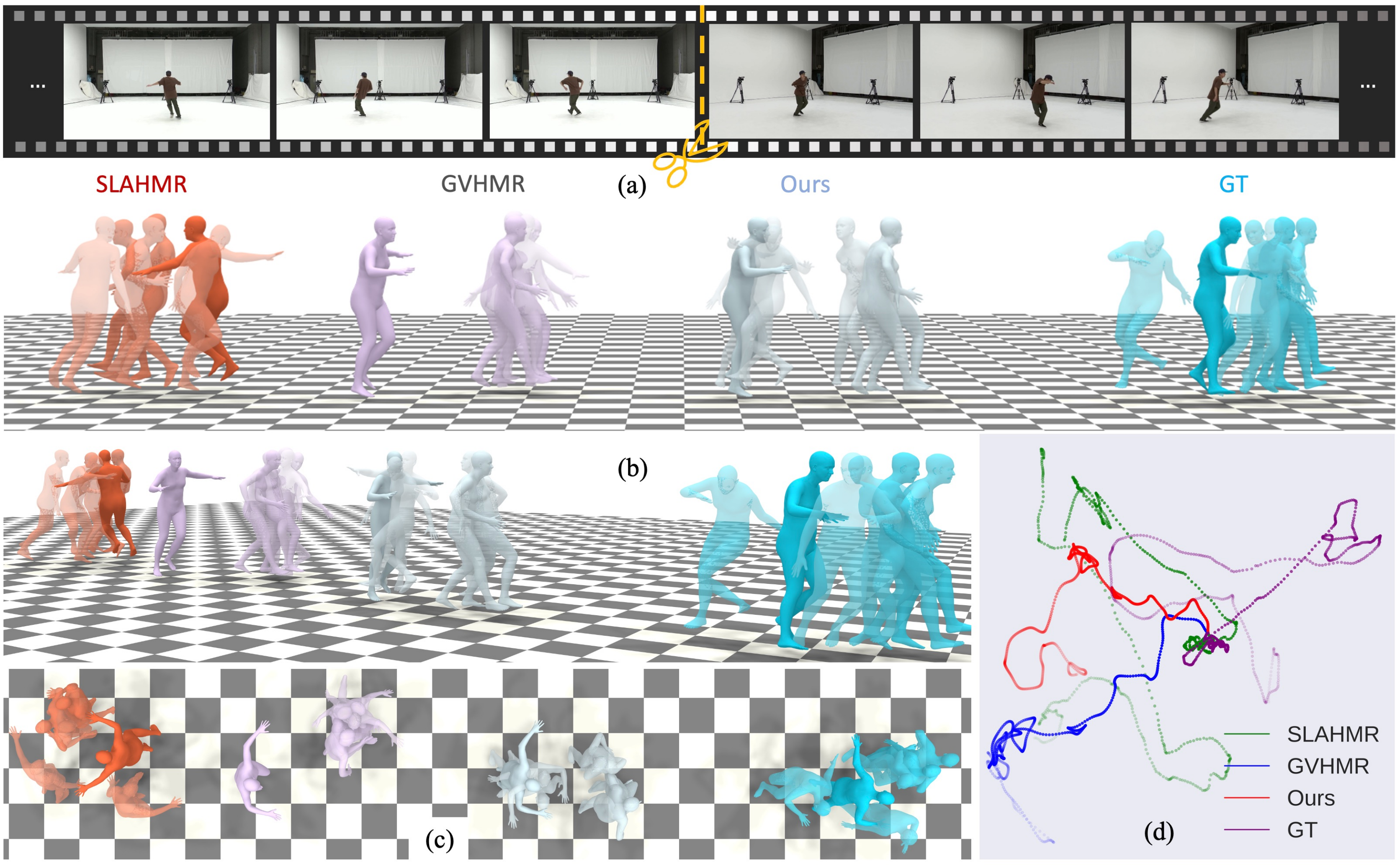}
    \label{fig:vis_comp_2}
\end{figure*}

\clearpage
\subsection{Visualization of in-the-wild multi-shot video}
To further validate the performance of our proposed method, we apply \methodname on a set of in-the-wild videos and provide the visualizations as shown in the figure. The frames at the top illustrate key moments in the video sequence, with the corresponding reconstructions visualized in 3D space below. The 3D reconstructions effectively capture the dynamic human poses and trajectories over time. In subfigure (a), the side view highlights the accurate reconstruction of intricate human motion, illustrating the consistency and smoothness of the trajectories even for fast and complex movements. Subfigure (b) presents an alternative side view, offering another perspective that underscores the spatial coherence of the reconstructions. Subfigure (c) provides a top-down view, offering an insightful perspective into the overall trajectory and positional accuracy of the reconstructed poses as they evolve over time. This view particularly emphasizes the spatial distribution and alignment of the poses in the reconstructed scene. These visualizations collectively demonstrate the robustness of \methodname\ in handling challenging, dynamic motions in diverse environments, further showcasing its applicability to real-world scenarios.

\begin{figure*}[htbp]
    \centering
    \includegraphics[width=0.98\linewidth]{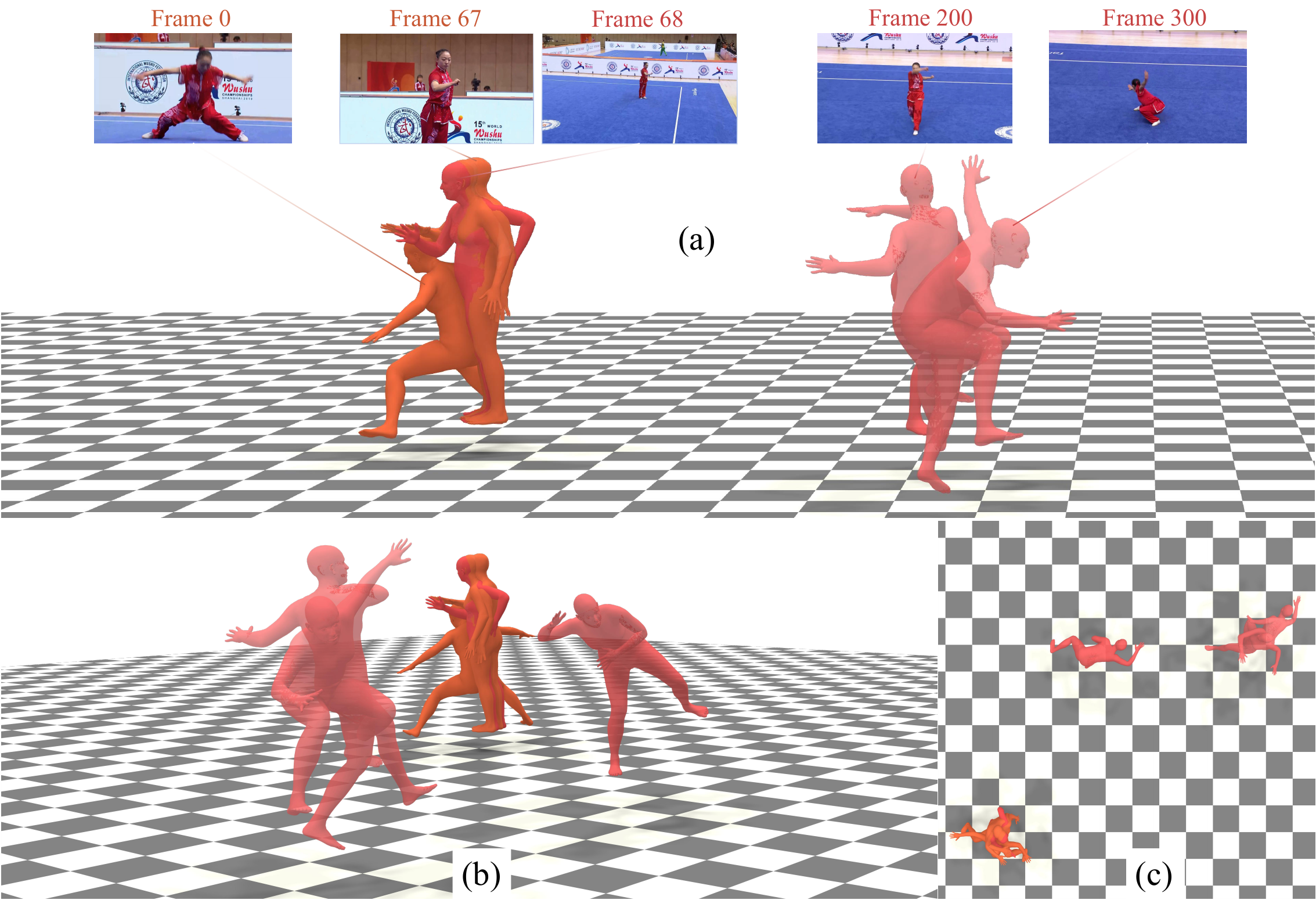}
    \label{fig:vis_comp_3}
\end{figure*}

\begin{figure*}[t]
    \centering
    \includegraphics[width=0.98\linewidth]{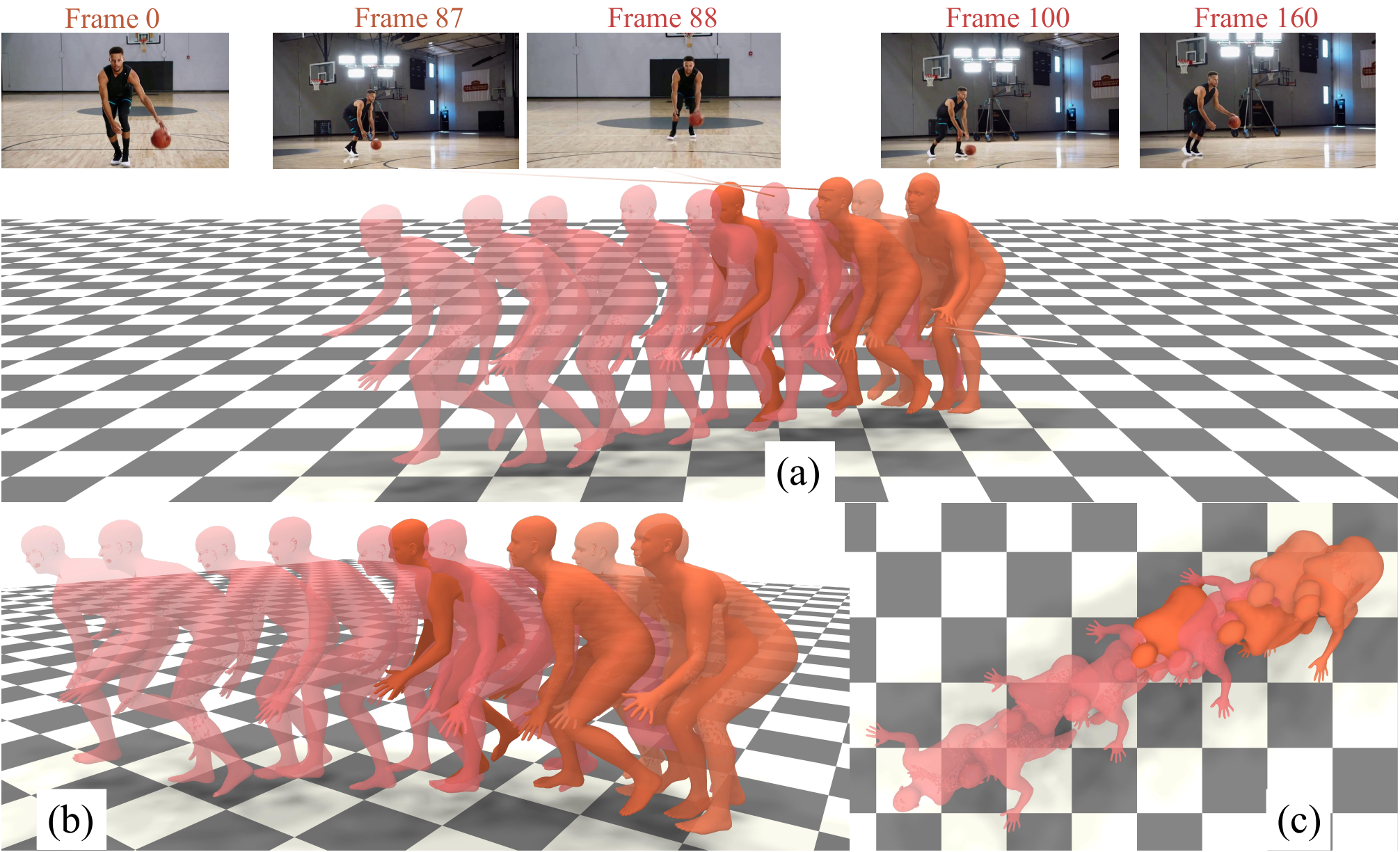}
    \label{fig:vis_comp_4}
\end{figure*}



\end{document}